\documentclass[letterpaper, 10pt, conference]{ieeeconf}
\usepackage[pdftex]{graphicx}

\usepackage{booktabs}
\usepackage{url}
\usepackage{amsmath,amssymb}
\usepackage{multicol,lipsum}

\usepackage{bm}
\usepackage{color, colortbl}
\usepackage[colorlinks,bookmarksopen,bookmarksnumbered,citecolor=red,urlcolor=red]{hyperref}

\hypersetup
{
	pdftitle = {},
	pdfauthor = {},
	pdfsubject = { },
	pdfkeywords = { },
	pdftoolbar = true,
	colorlinks = true,
	linkcolor = black,
	citecolor = black,
	urlcolor = black,
}

\usepackage{xcolor}
\definecolor{blue_iit}{RGB}{51,51,255}
\usepackage{algpseudocode}
\usepackage{algorithm}
\usepackage{glossaries}
\usepackage[tight]{units}
\usepackage[normalem]{ulem} 
\usepackage{cancel}
\definecolor{Gray}{gray}{0.9}

\usepackage{subcaption}
\usepackage{tikz}
\usepackage{scalerel}
\newcommand{\startmarker}{\scalerel*{
\begin{tikzpicture}[baseline=-0.5ex]
    \useasboundingbox (0,0) rectangle (0.3,0.3);
    \filldraw[fill=green, draw=black, line width=0.8pt]
        (0,0) -- (0.15,0.25) -- (0.3,0) -- cycle;
\end{tikzpicture}
}{\bigtriangleup}}

\newcommand{\goalmarker}{\scalerel*{
\begin{tikzpicture}[baseline=-0.5ex]
    \useasboundingbox (0,0) rectangle (0.3,0.3);
    \draw[black, line width=4.8pt] (0,0) -- (0.3,0.3);
    \draw[black, line width=4.8pt] (0.3,0) -- (0,0.3);
    \draw[red, line width=3.2pt] (0,0) -- (0.3,0.3);
    \draw[red, line width=3.2pt] (0.3,0) -- (0,0.3);
\end{tikzpicture}
}{\times}}

\newcommand{\achievedmarker}{
\begin{tikzpicture}[baseline={(0.15,0.)}]
    \useasboundingbox (0,0) rectangle (0.3,0.3);
    \filldraw[fill=yellow, draw=black, line width=0.8pt]
        (0.15,0.30)
        -- (0.18,0.20)
        -- (0.30,0.20)
        -- (0.20,0.13)
        -- (0.24,0.00)
        -- (0.15,0.08)
        -- (0.06,0.00)
        -- (0.10,0.13)
        -- (0.00,0.20)
        -- (0.12,0.20)
        -- cycle;
\end{tikzpicture}
}

\newcommand{\jumpmarker}{
\begin{tikzpicture}[baseline=-0.5ex]
    \filldraw[fill=yellow, draw=black, line width=0.8pt]
        (0,0) circle (0.1cm); 
\end{tikzpicture}
}

\IEEEoverridecommandlockouts                              

\usepackage[
  backend=biber,
  style=numeric,
  sorting=none,
  giveninits=true,
  maxbibnames=1,
  minbibnames=1
]{biblatex}

\AtEveryBibitem{%
  \clearfield{volume}
  \clearfield{number}
  \clearfield{pages}
  \clearfield{doi}
  \clearfield{url}
  \clearfield{isbn}
  \clearfield{issn}
  \clearfield{month}
  \clearlist{publisher}
  \clearlist{location}
  \clearfield{note}
}

\DeclareBibliographyDriver{article}{%
  \printnames{author}%
  \setunit{\addcomma\space}
  \printfield{title}%
  \setunit{\addcomma\space}
  \printfield{journaltitle}%
  \setunit{\addcomma\space}
  \printfield{year}%
  \finentry
}

\DeclareBibliographyDriver{inproceedings}{%
  \printnames{author}%
  \setunit{\addcomma\space}
  \printfield{title}%
  \setunit{\addcomma\space}
  \printfield{booktitle}%
  \setunit{\addcomma\space}
  \printfield{year}%
  \finentry
}

\newacronym{lf}{LF}{Left-Front}
\newacronym{rf}{RF}{Right-Front}
\newacronym{lh}{LH}{Left-Hind}
\newacronym{rh}{RH}{Right-Hind}

\newacronym{haa}{HAA}{Hip Adduction-Abduction}
\newacronym{hfe}{HFE}{Hip Flexion-Extension}
\newacronym{kfe}{KFE}{Knee Flexion-Extension}

\newacronym{imu}{IMU}{Inertial Measurement Unit}
\newacronym{dofs}{DoFs}{Degrees of Freedom}
\newacronym{rt}{RT}{Real Time}

\newacronym{urdf}{URDF}{Unified Robot Description Format}

\newacronym{com}{CoM}{Center of Mass}
\newacronym{cop}{CoP}{Center of Pressure}
\newacronym{zmp}{ZMP}{Zero Moment Point}
\newacronym{icp}{ICP}{Instantaneous Capture Point}
\newacronym{cmp}{CMP}{Centroidal Moment Pivot}
\newacronym{grfs}{GRFs}{Ground Reaction Forces}

\newacronym{ls}{LS}{Least Square}
\newacronym{lp}{LP}{Linear Program}

\newacronym{slip}{SLIP}{Spring Loaded Inverted Pendulum}
\newacronym{eom}{EoM}{Equation of Motions}
\newacronym{qp}{QP}{Quadratic Program}
\newacronym{sqp}{SQP}{Sequential Quadratic Programming}
\newacronym{mic}{MIC}{Mixed-Integer Convex}
\newacronym{cmaes}{CMA-ES}{Covariance Matrix Adaptation Evolution Strategy}
\newacronym{ara}{ARA*}{Anytime Repairing A*}
\newacronym{pca}{PCA}{Principal Component Analysis}
\newacronym{cpg}{CPG}{Central Pattern Generator}
\newacronym{wbc}{WBC}{Whole-Body Control}
\newacronym{cwc}{CWC}{Contact Wrench Cone}
\newacronym{fwp}{FWP}{Feasible Wrench Polytope}
\newacronym{nlp}{NLP}{Non Linear Program}
\newacronym{ocp}{OCP}{Optimal Control Problem}
\newacronym{pd}{PD}{Proportional-Derivative}

\newacronym{mpc}{MPC}{Model Predictive Control}
\newacronym{nmpc}{NMPC}{Nonlinear Model Predictive Control}

\newacronym{stance}{STANCE}{\textbf{S}oft \textbf{T}errain \textbf{A}daptation a\textbf{n}d \textbf{C}ompliance \textbf{E}stimation}

\newacronym{wbopt}{WBOpt}{Whole Body Optimization}

\newacronym{hc}{HC}{Hunt and Crossley's}
\newacronym{kv}{KV}{Kelvin-Voigt's}

\newacronym{wllsr}{WLLSR}{Weighted Linear Least Squared Regression}

\newacronym{mae}{MAE}{Mean Absolute Tracking Error}

\newacronym{ode}{ODE}{Open Dynamics Engine}
\newacronym{lip}{LIP}{Linear Inverted Pendulum}
\newacronym{srbd}{SRBD}{Single Rigid Body Dynamics}

\newacronym{cem}{CEM}{Cross Entropy Method}

\newcommand{\Rnum}{\mathbb{R}} 

\newcommand{\vect}[1]{\mathbf{#1}} 

\newcommand{\mat}[1]{\ensuremath{\begin{bmatrix}#1\end{bmatrix}}}	

\usepackage{xcolor}                    
\usepackage{tikz}                      
\usetikzlibrary{shapes.geometric}      

\DeclareRobustCommand{\startmarker}{%
  \tikz[baseline=-0.6ex]{%
    \node[regular polygon, regular polygon sides=3,
          inner sep=0pt, minimum size=1.6ex,
          draw=black, fill=green!60!black] {};%
  }%
}

\DeclareRobustCommand{\goalmarker}{%
  \tikz[baseline=-0.6ex]{%
    \draw[line width=0.9pt, red]
      (-0.65ex,-0.65ex) -- (0.65ex,0.65ex)
      (-0.65ex,0.65ex) -- (0.65ex,-0.65ex);
  }%
}

\DeclareRobustCommand{\jumpmarker}{%
  \tikz[baseline=-0.6ex]{%
    \node[circle, inner sep=0pt, minimum size=1.6ex,
          draw=black, fill=yellow!85!black] {};%
  }%
}

\DeclareRobustCommand{\achievedmarker}{%
  \tikz[baseline=-0.6ex]{%
    \node[star, star points=5, inner sep=0pt, minimum size=1.8ex,
          draw=black, fill=yellow!85!black] {};%
  }%
}

\newcommand\BibTeX{{\rmfamily B\kern-.05em \textsc{i\kern-.025em b}\kern-.08em
T\kern-.1667em\lower.7ex\hbox{E}\kern-.125emX}}

\makeatletter
\newcounter{definition*}
\newenvironment{definition*}[1][htb]
{\renewcommand{\ALG@name}{Definition}
	\let\c@algocf\c@megaalgorithm
	\begin{algorithm*}[#1]%
	}{\end{algorithm*}}
\makeatother

\makeatletter
\newcounter{definition}

\makeatother

\definecolor{sfahmi_blue}{RGB}{0.19,0.51,0.74}
\definecolor{LightBlue}{RGB}{0.4,0.4,1}

 \newcommand{\leg}{\mathrm{leg}}

\algnewcommand\algorithmicforeach{\textbf{for each}}
\algdef{S}[FOR]{ForEach}[1]{\algorithmicforeach\  #1\ \algorithmicdo}

\title{\LARGE \bf Bi-Level Optimization for Contact and Motion Planning in Rope-Assisted Legged Robots}%
\author{ Ruben Malacarne$^{1}$, Ioannis Tsikelis$^{2}$, Enrico Mingo Hoffman$^{2}$, and Michele Focchi$^{1}$ 
\thanks{$^1$ The authors are with the Dipartimento di Ingegneria e Scienza dell'Informazione (DISI), University of Trento. Email:  \href{mailto:name.surname@unitn.it}{name.surname@unitn.it}}
\thanks{$^2$ The authors are with Inria, Universit\'{e} de Lorraine, CNRS, 54000 Nancy, France.  Email:  \href{mailto:ioannis.tsikelis@inria.fr}{ioannis.tsikelis@inria.fr}, \href{mailto:enrico.mingo-hoffman@inria.fr}{enrico.mingo-hoffman@inria.fr}.}%
\thanks{The publication was created with the co-financing of the European Union FSE-REACT-EU, PON Research and Innovation 2014-2020 DM1062 / 2021, the French National Research Agency (ANR) under the project ANR-24-CE33-0753-01 (MeRLin), and of the VRT foundation under the project ``ALPINE: Robot intelligente e autonomo per operazioni in ambienti montani''.} }

\begin{document}
\maketitle

\begin{abstract}
This paper presents a planning pipeline framework for locomotion in rope-assisted robots climbing vertical surfaces. 
The proposed framework is formulated as a bi-level optimization scheme that addresses a mixed-integer problem: selecting feasible terrain regions for landing while simultaneously optimizing the control inputs, namely rope tensions and leg forces, and landing location.
The outer level of the optimization is solved using the Cross-Entropy Method, while the inner level relies on gradient-based nonlinear optimization to compute dynamically feasible motions. 
The approach is validated on a novel climbing robot platform, \emph{ALPINE}, across a variety of challenging terrain configurations.
%
%
\end{abstract}


\section{Introduction and Related Works}\label{sec:introduction}
Robotic systems operating in mountainous environments face several challenges arising from the harshness of these settings, including severe climatic conditions and highly variable morphology~\cite{marconi2012sherpa, miki2022learning}.
%
%
Climbing robots, in particular, have been employed to navigate uneven terrain, with some systems adopting alternative locomotion strategies, such as ropes, to ascend extremely steep or vertical surfaces~\cite{uckert2020investigating, alpine, cepolina2006roboclimber}.
However, while ropes enable access to otherwise inaccessible spaces, they increase the complexity of planning and control strategies due to several factors, most notably the \emph{underactuation} arising from the unilateral nature of rope tension under gravity.
\par
Despite several works in the literature addressing motion and contact planning for climbing legged robots~\cite{bretl2006motion}, only a limited number explicitly incorporate ropes and their effects.
For example, in~\cite{hoffman21}, optimal control is employed to plan motions and contacts for dynamic rappelling for a legged system anchored by a single rope, using a template model. 
In~\cite{choi2023examining}, the effects of rope bending on climbing systems are examined.
\cite{alpine} addresses the generation of a feasible pendular motion for a novel platform named \emph{ALPINE}, combining the action of a leg and adjustable ropes to connect a starting location to a landing location. 
\par
Concerning contact planning strategies, one class comprises contact-implicit methods~\cite{Mordatch2012DiscoveryOC, kim2025contact-implicit}, which model the robot dynamics as a hybrid dynamical system. 
To make the resulting optimization problem computationally tractable, these approaches typically rely on relaxations of the complementarity constraints~\cite{Manchester24}.
A second class of methods explicitly parameterizes contact locations within the optimization, yielding mixed-integer nonlinear programs~\cite{Cauligi20}. 
While this formulation enables direct reasoning over discrete contact decisions, its computational complexity scales exponentially with the number of candidate contacts.
\begin{figure}[!tbh]
    \centering
    \includegraphics[width=0.45\columnwidth]{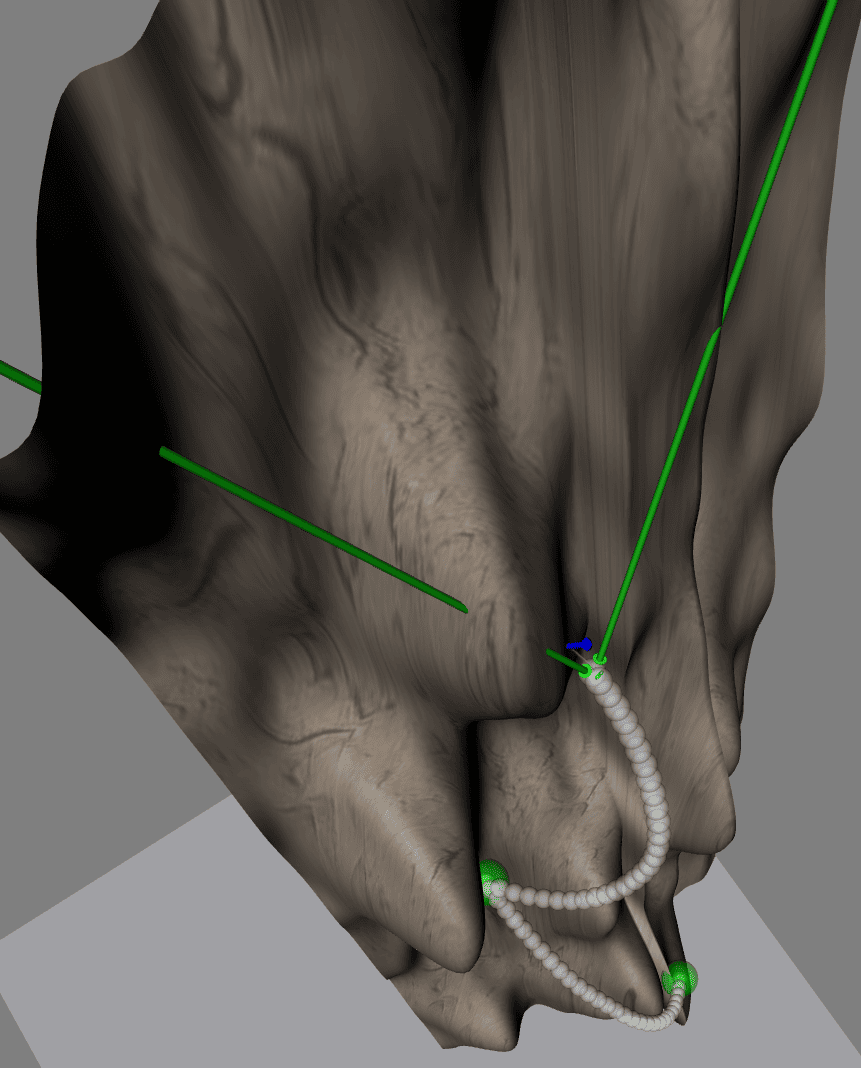}
    \includegraphics[width=0.45\columnwidth]{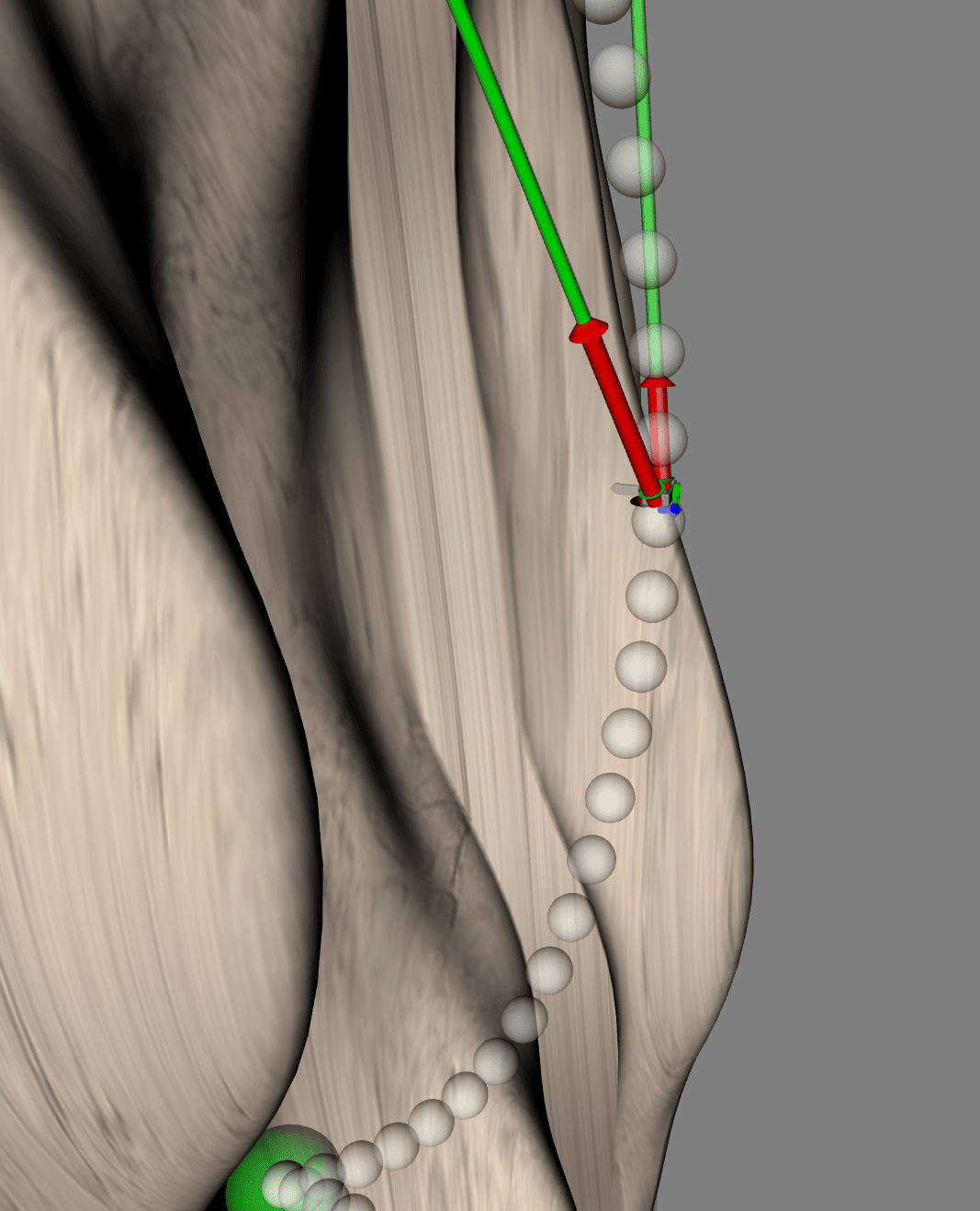}
    \caption{\small (left) Dynamic simulation of the \emph{ALPINE} robot performing a multi-jump maneuver on a rough wall; (right) close-up of the robot during the first intermediate (downward) jump. The reference trajectory is shown in white, while the green marker indicates the intermediate landing target. The ropes are depicted in green, and the red arrows represent the instantaneous pulling forces.}
    \label{fig:gazebo_simulation}
\end{figure}

Hybrid methods seek to bridge these two approaches by pre-selecting a subset of gait or contact parameters while optimizing the remaining variables. 
A particularly promising direction is the use of a bi-level hybrid optimization framework that combines black-box optimization with optimization-based motion planning, enabling the independent discovery of feasible contact sequences for a given motion task~\cite{Tsikelis25,dhedin2025simultaneous}.
%
%
%
Hence, despite the extensive existing literature on motion and contact planning, the same problem, when formulated for rope-assisted robots under physical and environmental constraints, has received comparatively little attention.
\par
\noindent \textbf{Contribution.} In this regard, this work presents a computationally efficient algorithm for planning sequences of jumping motions for climbing robots assisted by ropes, explicitly accounting for both the physical limitations of the hardware and constraints imposed by the environment.
We formulate the problem of simultaneous contact and motion planning as a bi-level nested optimization, in both discrete and continuous decision variables:
\begin{itemize}
    \item the outer loop consists of a black-box optimizer that handles contact scheduling (number of jumps) and patch selection by dividing the environment into discrete patches;
    \item the inner loop consists of a gradient-based optimization that attempts to provide a feasible jumping trajectory and contact placements within the selected patches.
\end{itemize}
%
%
%
%
%
The efficacy of the proposed planning strategy is tested on multiple terrains and in dynamic simulations involving the \emph{ALPINE} robot, presented in~\cite{alpine}.
It is worth noticing that the proposed method aligns with the work in~\cite{Marcucci24}, with the graph search substituted by a black-box optimizer.
\par
The paper is organized as follows: Section~\ref{sec:preliminaries} introduces the particular \emph{ALPINE} robot kinematics and dynamics model, as well as the cost map used to define the environment patches; Section~\ref{sec:methodology} presents the bi-level optimization strategy and its components; Section~\ref{sec:results} showcases planning and simulation experiments; and finally, Section~\ref{sec:conclusion} provides conclusions and discusses future work.


\section{Preliminaries}\label{sec:preliminaries}
\subsection{ALPINE robot platform} \label{subsec:alpine}
\begin{figure*}[!tb]
    \centering
\begin{subfigure}{0.33\textwidth}
    \centering
    \includegraphics[width=\linewidth, trim={2cm 1.5cm 2cm 2cm}, clip=true]{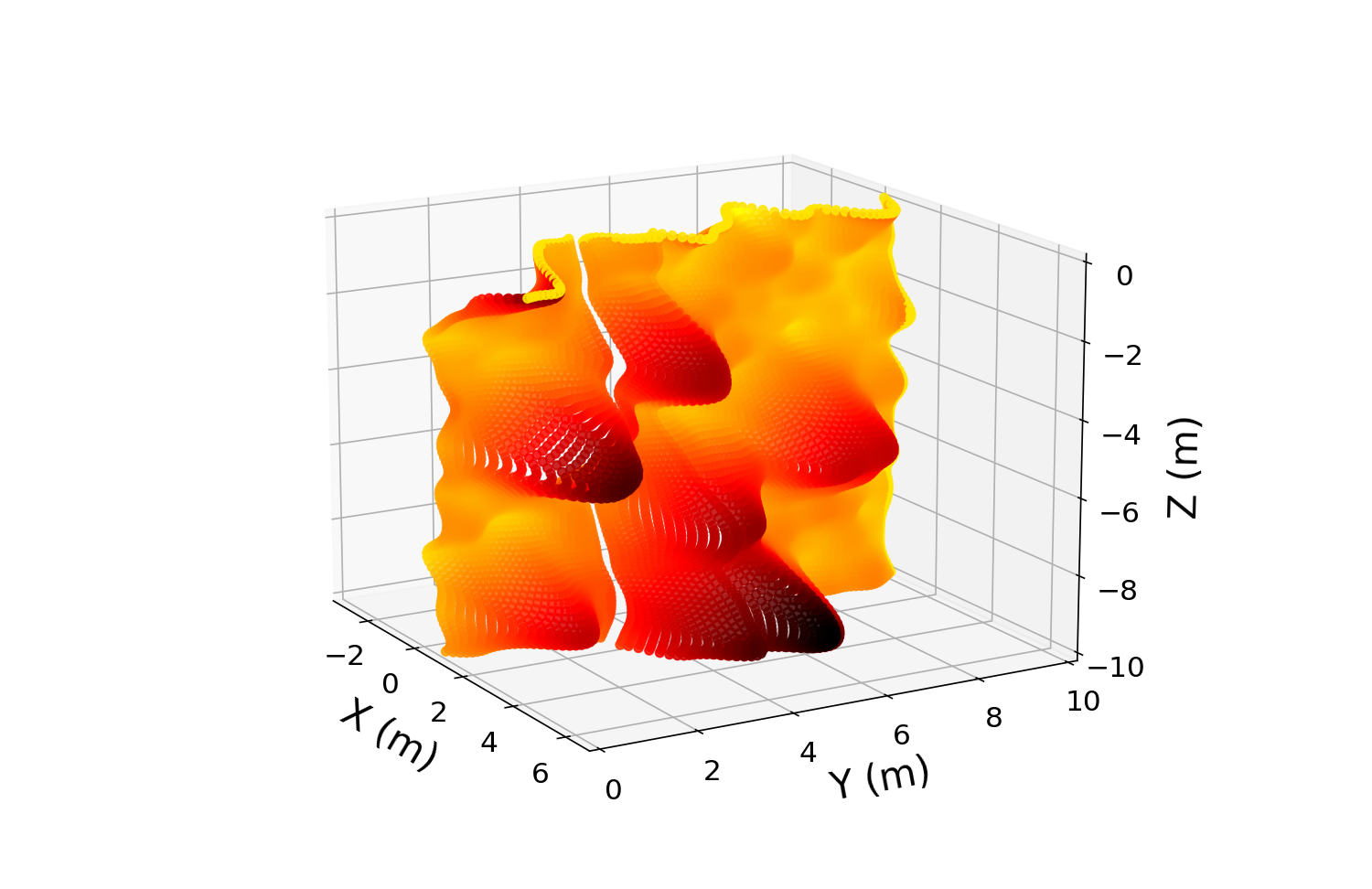}
    \caption{Height map.}
\end{subfigure}\hfill
\begin{subfigure}{0.33\textwidth}
    \centering
    \includegraphics[width=\linewidth, trim={2cm 1.5cm 2cm 2cm}, clip=true]{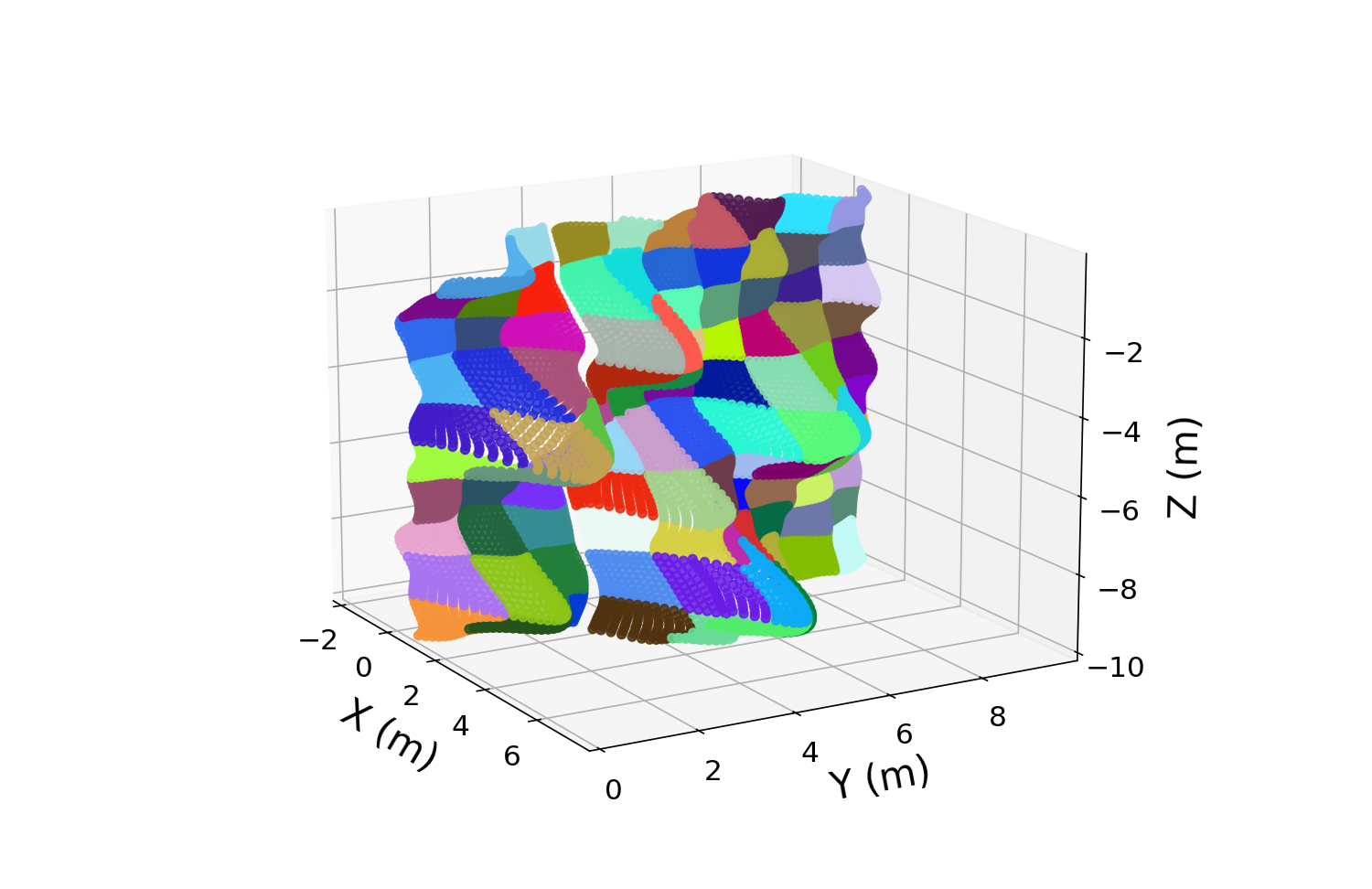}
    \caption{Patch division.}
\end{subfigure}\hfill
\begin{subfigure}{0.33\textwidth}
    \centering
    \includegraphics[width=\linewidth, trim={2cm 1.5cm 2cm 2cm}, clip=true]{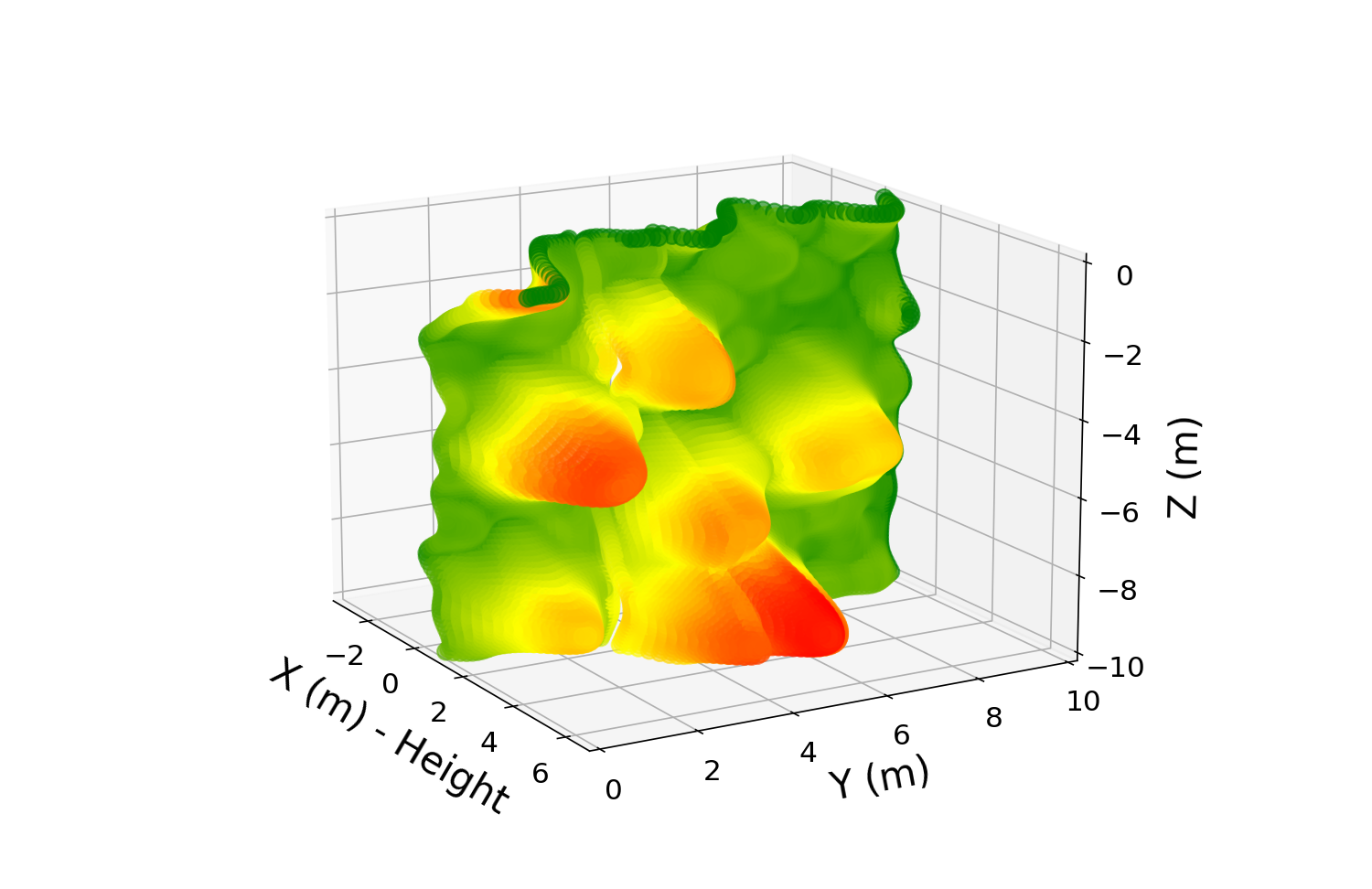}
    \caption{Cost map.}
\end{subfigure}
    \caption{\small The Cost map is generated from the height map of a rocky wall. The color represents the slope intensity at $10 \ cm$ resolution. The cost values are represented using a color scale, with green indicating the minimum cost and red the maximum.}
    \label{fig:full_cost_map}
    \vspace{-0.5cm}
\end{figure*}

\begin{figure}[tbh!]
\begin{subfigure}{1\columnwidth}
  \centering
  \includegraphics[width=0.6\columnwidth, trim={0cm 0cm 0cm 3cm}, clip=true]{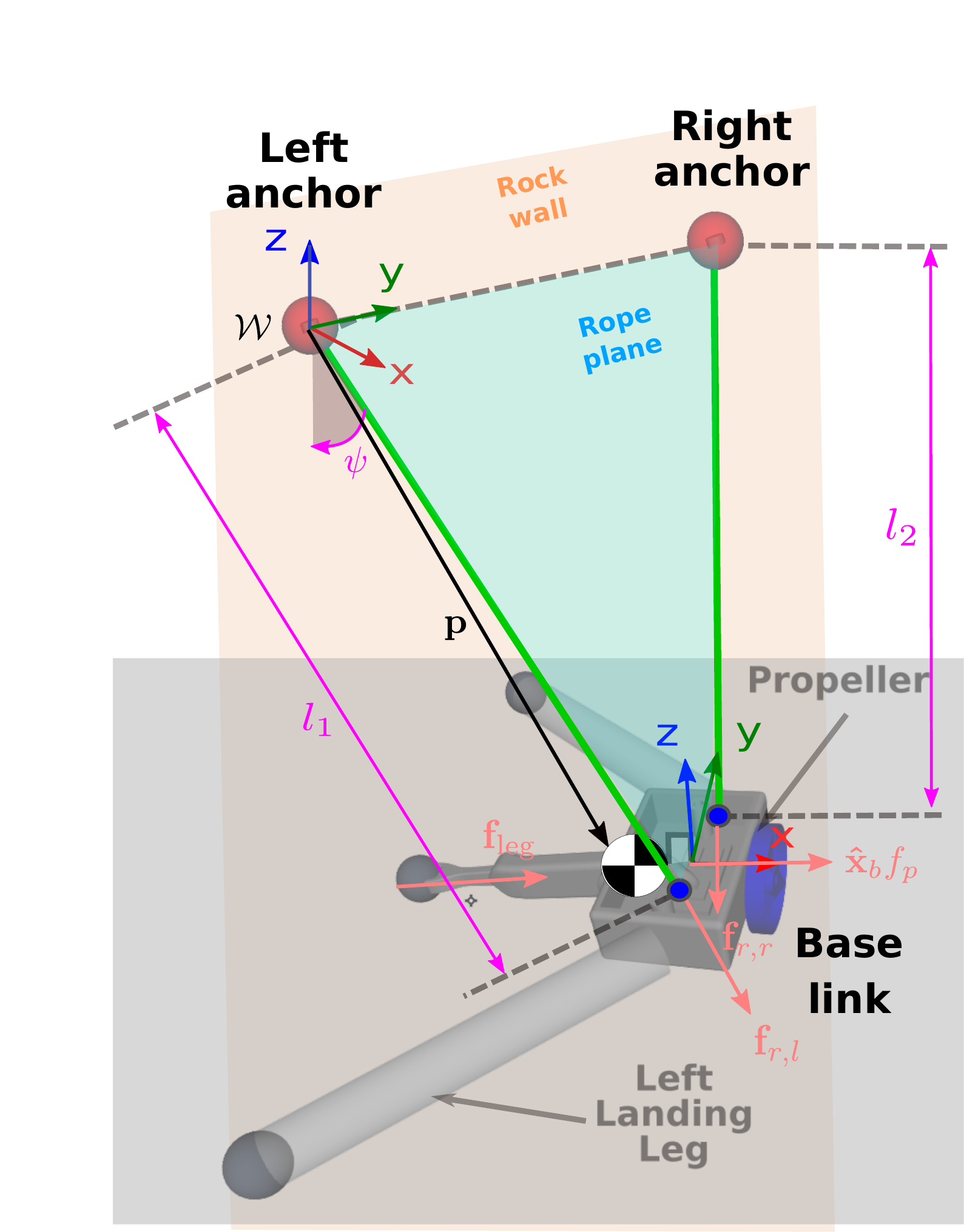}
  \caption{Kinematic model of the \emph{ALPINE} robot. A prismatic leg initiates the jump. Two ropes guide the motion during flight, while a rear-mounted propeller improves tracking performance. Two lateral legs attached to the base help stabilize the landing. The inertial frame $\mathcal{W}$ is attached to the left anchor frame. The actuation forces are shown in light red, while the generalized coordinates of the reduced-order model are depicted in pink.}
  \label{fig:3dmodel2anchors_propellers}
\end{subfigure}
\vspace{0.5cm}
\\
\begin{subfigure}{1\columnwidth}
  \centering
  \includegraphics[width=0.6\columnwidth]{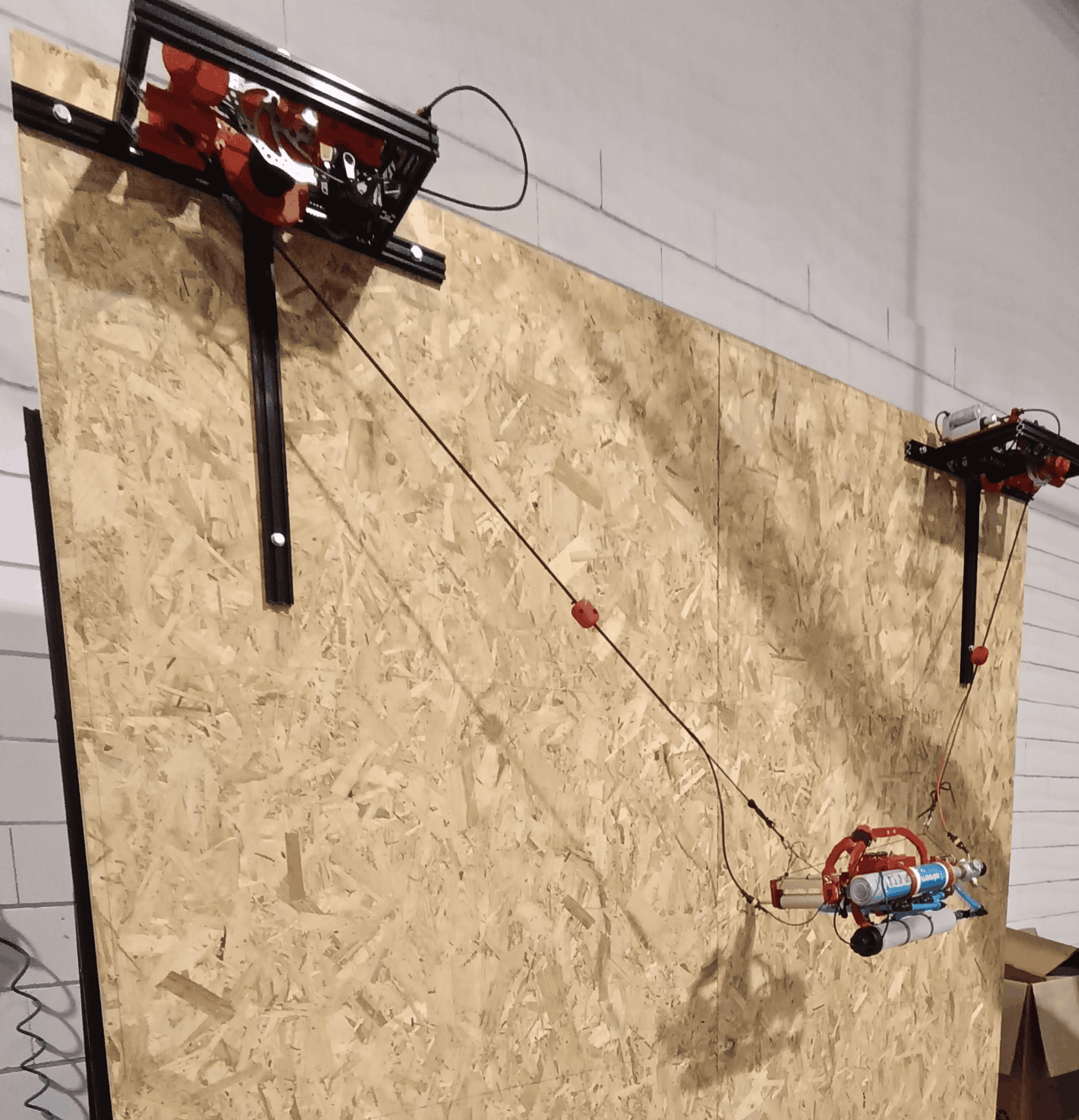}
  \caption{Built prototype of the ALPINE robot.}
  \label{fig:built_prototype}
\end{subfigure}
\caption{\small The \emph{ALPINE} robot. On top, the kinematics model, and on the bottom, the real prototype.}
\end{figure}

\label{sec:alpine}
A robot hanging on a single rope~\cite{focchi23icra,  hoffman21} has severe limitations in terms of possible lateral motions, and hence reachable locations. 
Additionally, the lateral pull of the rope makes the system unstable when static, limiting its ability to execute tasks. 
The \emph{ALPINE} robot~\cite{alpine} overcomes these limitations, considering two ropes attached to  \textit{fixed} anchors on the wall. 
Both ropes can be \textit{independently} wound or unwound via hoist motors (see Fig.~\ref{fig:3dmodel2anchors_propellers} and Fig.~\ref{fig:built_prototype}). 
A prismatic leg initiates the jumping motion by rapidly pushing the robot away from the slope. 
During the flight phase, the ropes are actively wound or unwound to regulate the robot’s motion during the flight, enabling it to swing over the wall surface and overcome obstacles.
%
%
During landing, the excess of kinetic energy is dissipated via two landing legs with passive torsional springs or via the prismatic leg itself. 
A propeller is mounted on the rear of the robot to reject tracking errors in the direction normal to the ropes' plane during the flight phase. 
%
%
The full detailed model for the \emph{ALPINE} robot is underactuated and requires accounting for a holonomic kinematic loop constraint.
However, by neglecting angular dynamics and assuming that the mass is concentrated at the \gls{com}, with the additional assumption that the ropes remain taut at all times, a reduced-order (3DOFs) model can be adopted. 
In this formulation, the holonomic constraint is eliminated by an appropriate choice of generalized coordinates, namely $\vect{q}_r =\mat{ \psi & l_1 & l_2} \in \Rnum^3$, where $\psi$ is the angle between the plane defined by the ropes and the vertical plane passing through anchors and $l_1$, $l_2$ are the lengths of the left and right ropes, respectively. 
The dynamics of the robot in an inertial frame $\mathcal{W}$ attached to the \textit{left} anchor point, is:
\begin{equation}
  m (\vect{\ddot{p}} - \vect{g}) = \underbrace{\vect{\hat{a}}_{rl}
    f_{rl}}_{\vect{f}_{rl}} + \underbrace{\vect{\hat{a}}_{rr}
    f_{rr}}_{\vect{f}_{rr}} + \vect{f}_{leg} 
  \label{eq:newton}
\end{equation}
where $\vect{\hat{a}}_{ri} = \frac{\vect{p} - \vect{p}_{ai}}{\Vert \vect{p} - \vect{p}_{ai} \Vert} \in \Rnum^3$ and $f_{ri} \in \Rnum$, with $i = \{r,l\}$, are the rope axes and 
the magnitude of the exerted forces, respectively, with the two anchor point positions defined by $\vect{p}_{al}$ and $\vect{p}_{ar}$, respectively\footnote{While in Section 
\ref{sec:gazebo_sim} we simulate the full model depicted in Fig. \ref{fig:3dmodel2anchors_propellers}, in the reduced model, for the sake of simplicity, we assume the rope 
attachment points to be coincident with the robot’s \gls{com}.}.
The expression of the forward kinematics for the position $\vect{p}$ of the robot (origin of the base-link) as a function of $\vect{q}_r$ is given by:  
\begin{equation}
  \vect{p}(\vect{q}_r) = \mat{p_x \\ p_y\\ p_z}=
  \mat{l_1 \sin(\psi) \sqrt{1 - \frac{(d_a^2 + l_1^2 - l_2^2)^2}{4 d_a^2 l_1^2}} \\	
    (d_a^2 + l_1^2 - l_2^2)/(2 d_a) \\
    -l_1 \cos(\psi) \sqrt{1 - \frac{(d_a^2 + l_1^2 - l_2^2)^2}{4 d_a^2l_1^2}}}.
  \label{eq:fwd_kin_minimal}
\end{equation}
\par
Double differentiating w.r.t the state variables, the final equation of the reduced model \eqref{eq:simplified_2ropes_minimal} can be obtained  (see \cite{alpine} for details) in terms of the state variables  where the holonomic constraint has been embedded (by construction) thanks to the specific choice of  coordinates $\vect{q}_r$:
\begin{equation}
  \mat{\ddot{\psi}\\ \ddot{l}_1\\ \ddot{l}_2} = \vect{A}_{d}^{-1}\left[ \frac{1}{m} \vect{f}_{tot} - \vect{b}_{d} \right],
\label{eq:simplified_2ropes_minimal}
\end{equation}
with $\vect{A}_{d}$ and $\vect{b}_{d}$ reported in the Appendix of~\cite{alpine}.
The reduced model was shown in \cite{alpine} to be a good approximation of the real system, and it is sufficient to generate feasible jump trajectories that do not involve large orientation changes.
\subsection{Cost Map}
\label{sec:cost_map}
Regarding the terrain (e.g. rock wall) on which the robot operates, the proposed planning framework relies on a comprehensive cost map derived from an acquired terrain \textit{height map}.
We assume that the input terrain point-cloud has undergone pre-processing, such as outlier rejection and inpainting, to remove spurious measurements and fill missing regions, thereby eliminating holes and ensuring a spatially consistent representation. 
The resulting sparse point cloud is then interpolated into a continuous surface representation. 
This step enables the application of convolution operations with multiple kernels, whose outputs are subsequently mapped to specific cost values at each point of the terrain.
In addition to the filters evaluating the surface morphology, we also  incorporate a distance-based cost relative to the robot’s anchoring points. 
This term penalizes terrain regions that lie significantly deeper with respect to the robot’s anchor location as they would lead to statically infeasible configurations for the robot (that cannot address overhangs).
The level of detail at which the terrain is divided into patches is  selected manually, based on the map dimensions.
%
%
For each candidate foothold, we compute a quality score defined as a linear combination of slope, and roughness metrics, obtained from the above-mentioned filters.
%
%
Each cell in the height map is associated with a landing cost $c_l(Y, Z)$ that represents the quality of the landing location and is a linear combination of different quality measures:
\begin{equation}
    c_l(Y, Z)=\sum_i w_{cm,i} c_i,
\end{equation}
where $\mathbf{w}_{cm}=\mat{w_{sl} & w_{sd} & w_{d}}$   
represents the weight vector assigned to the morphological filters and quality metrics:
\begin{itemize}
    \item \textbf{$w_{sl}$ (Slope filter):} filter associated with first-order derivatives, to penalize the slope.
    \item \textbf{$w_{sd}$ (Roughness filter):} filter associated with second-order derivatives to penalize abrupt changes in curvature.
    \item \textbf{$w_{d}$ (Deep filter):} heavily penalizes regions inside the anchors' vertical.
\end{itemize}
%
%
%
Fig. \ref{fig:full_cost_map} shows the computed cost map generated from a height map of the rock wall. 
%

\section{Methodology}\label{sec:methodology}
\begin{figure}[!htb]
    \centering
    \includegraphics[height=0.7\columnwidth]{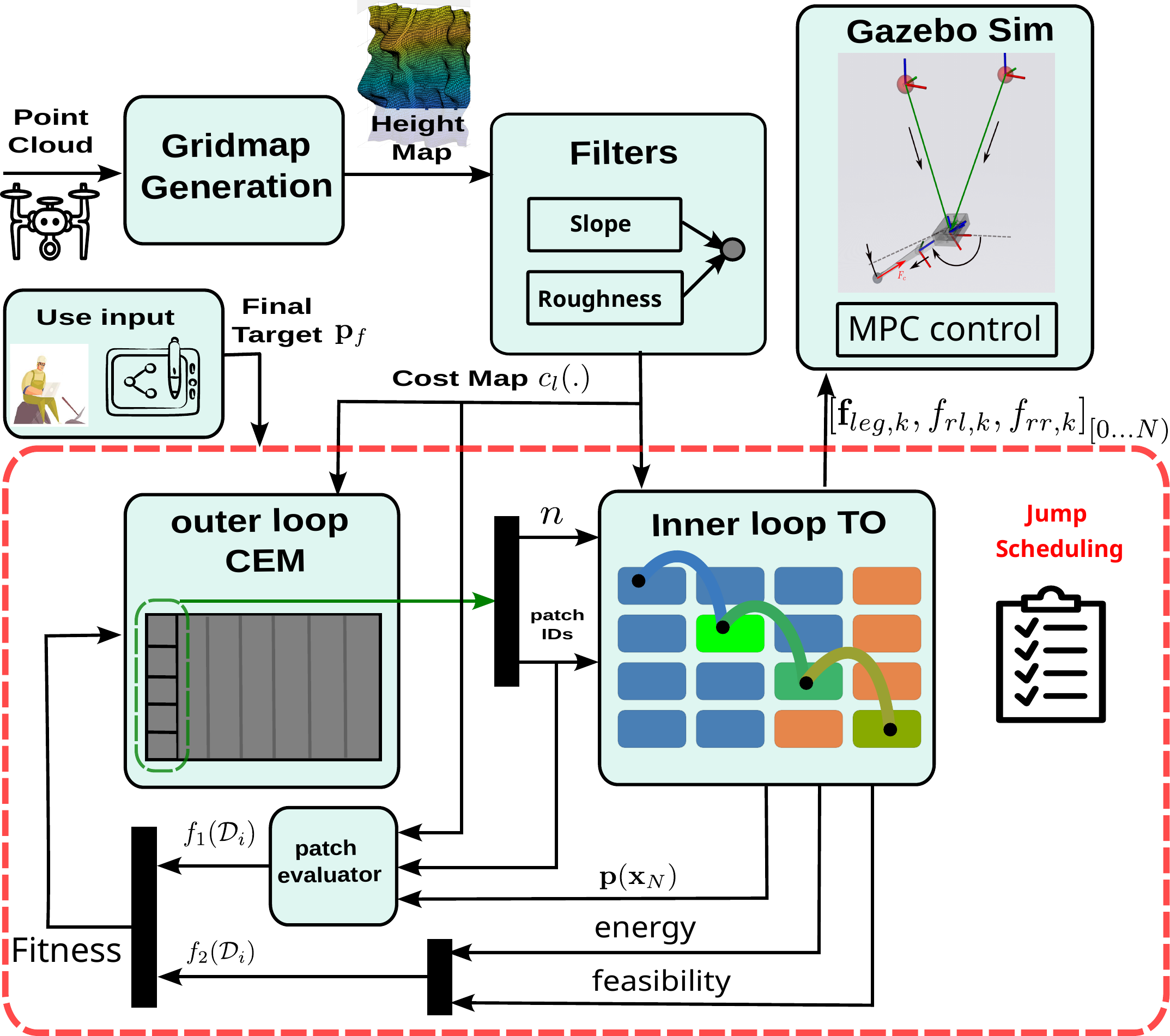}
    \caption{\small Block Diagram of the bi-level multi-jump optimization pipeline for the \emph{ALPINE} robot.}
    \label{fig:block_diagram}
    \vspace{-0.25cm}
\end{figure}

The proposed planning algorithm is based on a bi-level optimization scheme. 
The \textit{outer loop} employs a gradient-free \gls{cem} method to handle integer variables, such as patch selection and the number of jumps.
The \textit{inner loop} performs nonlinear trajectory optimization to compute a set of trajectories corresponding to the number of jumps selected by the outer optimization, while respecting under-actuation, hardware, and rope constraints, and simultaneously determining the optimal placement of landing contacts within the selected patches.
A block diagram of the pipeline is shown in Fig. \ref{fig:block_diagram}.
This formulation enables the inner loop to plan physically consistent motions and landing locations, while the outer loop selects a minimal sequence of intermediate jumps and landing patches to connect start and goal, based on energy consumption, inner-loop convergence quality, and terrain cost information.
\par
We denote by $\mathcal{D}$ the set of integer decision variables, namely the number of jumps $n \in \mathbb{Z}^+, \ n \ge 1$ and the indices of the selected intermediate patches $\boldsymbol{\pi} \in \mathbb{Z}^{n-1}$:
\begin{equation}
    \mathcal{D} = \left\{n, \boldsymbol{\pi}\right\}.
    \label{eq:integer_variables}
\end{equation}
Notice that the dimension of $\boldsymbol{\pi}$ is $n-1$ since the goal location is assumed to be known.
The bi-level optimization problem solved at each \emph{iteration} has the following form:
\begin{equation}
\begin{alignedat}{2}
\min_{\mathcal{D}} \; \mathcal{F}(\mathcal{D})
  &= \left\{ f(\mathcal{D}_0), f(\mathcal{D}_1), \dots, f(\mathcal{D}_{I-1}) \right\}& \\
\text{s.t.} \quad n_i &\ge 1, & \hspace{-1cm}\forall \ i \in I,\\
  \boldsymbol{\pi}_i &\in \boldsymbol{\Pi}_{ID}, & \hspace{-1cm}\forall \ i \in I, 
\end{alignedat}
\label{eq:minlp}
\end{equation}
where $I$ represents the number of \emph{individuals} of the \emph{population} for each \emph{iteration}, and $\mathcal{D}_{i}$ is a set of discrete variables associated with the individual $i$, composed by $n_i$  the number of jumps and $\boldsymbol{\Pi}_{ID}$ the set of all patches identifiers (IDs). 
The objective is to minimize the cost function $\mathcal{F}(\mathcal{D})$ by selecting the optimal number of jumps and a set of patch IDs from the catalog $\boldsymbol{\Pi}_{ID}$.
The cost $f(\mathcal{D}_i)$ is formed by summing 2 terms: 
\begin{equation}
    f(\mathcal{D}_i) = f_1(\mathcal{D}_i) + f_2(\mathcal{D}_i).
    \label{eq:fitness_sum}
\end{equation}
\par
The first term is the sum of the average terrain costs $\bar{c}_{t,j}$  associated with the selected patch $j$:
\begin{equation}
    f_{1}(\mathcal{D}_i) = w_{ac} \sum_{j\in\boldsymbol{\pi}_i}\bar{c}_{t,j}. 
    \label{eq:fitness_term_1}
\end{equation}
%
This enables us to rule out bad rock features like cracks, asperities,  holes, etc., favoring locally flat surfaces. 
\par
The second term results from the resolution of multiple non-linear \gls{ocp} parameterized with the set of integer variables $\mathcal{D}_i$ selected by the outer optimization, which will be detailed in Section~\ref{sec:inner_optim}. In particular, $f_2(\mathcal{D}_i)$ is composed of the cumulative consumed energy $c_e$  (with weights $w_e$) to jump between the selected patches. In case a constraint violation and/or non-convergence, for each sub jump, occurs,  a high penalty $w_p$ is returned.
\begin{equation}
    f_{2}(\mathcal{D}_i) = 
    \begin{cases} 
      \sum_{j=1}^{n} w_e c_{e,j} , & \text{if } \forall j \ \text{OCP}_j \text{ converges}, \\
      w_{p}, & \text{if } \exists j  : \text{OCP}_j  \text{ does not conv.}.
    \end{cases}
    \label{eq:fitness_term_2}
\end{equation}
Time is not considered as a feature to minimize because it is linked to the energy consumption, and the optimization might leverage a longer jump to accommodate a tighter constraint in the rope force magnitude. Consider, for instance, an upward jump: the limited actuation capability of the rope strongly constrains the achievable upward acceleration, thereby requiring a stronger leg push. This increased impulse allows the robot to remain airborne longer and thus reach the desired vertical displacement.
Whenever a loop is detected in the patch selection, i.e., when the same patch is visited two or more times within a single jump sequence, the cost $f_{2}(\mathcal{D}_i)$ is assigned a large penalty without executing the inner-loop optimization. 
This avoids unnecessary computation and reduces the likelihood of such candidate solutions being favored during sampling.
%
%
%
%
%
%
\subsection{Outer optimization: Cross Entropy Method}
\label{sec:outer_loop}
The outer loop of our bi-level optimization framework adopts the \gls{cem} due to its ability to handle integer variables and arbitrary, non-differentiable cost functions, and to converge toward a probability distribution whose mass is concentrated in regions of near-optimal solutions~\cite{rubinstein1999cross}.
\begin{algorithm}[t!]
    \caption{Cross-Entropy Method For Contact Selection}\label{algo:cem}
    \begin{algorithmic}[1]
        \Procedure{CEM}{$N_{\text{iter}}, N_{\text{elites}}, K, \mathcal{F}$}
            \State Initialize probability $\boldsymbol{r}^{[1]}$ 
            \For {$\text{iter}=1\to N_{\text{iter}}$}
                \State $\mathcal{D} = \{\mathcal{D}_{1}, \dots, \mathcal{D}_{_I}\}$, $\mathcal{D}_{i}\sim Cat(K, \boldsymbol{r}^{[k]})$\label{algo:sample}
                \State$\mathcal{F}(\mathcal{D}) = \{f(\mathcal{D}_{_1}), \dots, f(\mathcal{D}_{_I})\}$\label{algo:evaluate}
                \State$\mathcal{D}^{\text{elite}} = \{\mathcal{D}_{1}^{\text{elite}}, \dots, \mathcal{D}_{N_{\text{elites}}}^{\text{elite}}\}$
                \For {$i=1\to K$}
                    \State$\boldsymbol{r}^{[\text{iter}+1]} = \frac{\text{count\_occurrences}(K_i, \mathcal{D}^{\text{elite}})}{N_{\text{elites}}}$\label{algo:update}
                \EndFor
            \EndFor
        \EndProcedure
    \end{algorithmic}
\end{algorithm}
\gls{cem}, as shown in Algorithm~\ref{algo:cem}, is a black-box optimization method that models the optimal solution as a parameterized probability distribution, in our case, a categorical distribution. At every iteration, it generates a sample of population members, i.e., number of jumps and patch IDs, (Alg. \algref{algo:cem}{algo:sample}) and evaluates their \textit{fitness score}, i.e., the cost given by Eqfire.~\eqref{eq:fitness_sum}, (Alg. \algref{algo:cem}{algo:evaluate}). Finally, the probability of each category $\boldsymbol{r}$ is updated as the arithmetic mean of the times it was selected by a predefined percentage of the best fitting (lowest costing) candidates, also known as the \textit{elites} (Alg. \algref{algo:cem}{algo:update}).
%
%
%
\subsection{Inner optimization}
\label{sec:inner_optim}
%
The outer-loop optimization determines the number of intermediate jumps and the associated set of patches on which the robot must land to reach the goal. 
The purpose of this section is to describe how to connect them with feasible jumping motion plans from one patch to the next, and to select/favor the landing location with the lowest cost within each patch.
\par
The first component of the outer optimization output vector specifies the number of jumps, $n \leq n_{\max}$. Since all individuals in the population must share the same dimensionality, only the first $n$ elements of the discrete decision vector are evaluated. These elements encode the patch IDs associated with the intermediate landings.
For each selected patch, its centroid $\mathbf{p}_{c,i} \in \mathbb{R}^2$ and the corresponding cost function $c_i(\cdot)$ are provided as inputs to the trajectory optimization of the associated intermediate jump. The jumps are solved sequentially in cascade: the lift-off location coincides with the start robot position for the first jump, and with the previous landing position for all subsequent jumps.
%
Planning motions for the \emph{ALPINE} robot is challenging because the robot’s dynamics are highly nonlinear and the leg thrust force is impulsive and friction-limited.
As a result, we formulated the robot’s motion planning as a nonlinear \gls{ocp} transcribed into a finite-dimensional \gls{nlp} where we discretized the rope forces along the horizon in $N$ knots equally spaced by $dt = t_f/N$ time intervals, $\vect{X} = \left[\mathbf{x}_0, \dots, \mathbf{x}_N \right]$ state variables, and $\vect{U} = \left[\mathbf{u}_0, \dots, \mathbf{u}_{N-1} \right]$ controls\footnote{Here, the state and control variables are associated with the sub-jump $j$ of individual $i$, and the cost and constraint functions are additionally parameterized by the integer decision variables $\mathcal{D}_i$. These dependencies are omitted to avoid notational clutter.}:
%
%
%
\begin{subequations}
\label{eq:NLP}
\begin{align}	
	 \min_{\vect{X}, \vect{U}, t_f} & \sum_{k=0}^{N-1} \ell \left(\mathbf{x}_k, \mathbf{u}_k\right)+\ell_{\mathrm{f}}\left(\mathbf{x}_N\right)   \label{eq:cost_discrete}\\
	\text { s.t. } & \mathbf{x}_0=\hat{\mathbf{x}}_0,																					                   \label{eq:bound_discrete}   \\
	              & \mathbf{x}_{k+1}=  int(\mathbf{x}_{k}, \dot{\mathbf{x}}_{k}), & \forall \ k \in \left[0, N\right),      				   \label{eq:dyn_discrete}   \\
	              & \vect{h}_k \left(\mathbf{x}_k, \mathbf{u}_k\right) \leq 0, & \forall \ k \in \left[0, N\right),											   \label{eq:path_discrete} \\
                  & \vect{g}_k \left(\mathbf{x}_k\right) \leq 0, & \forall \ k \in \left[0, N\right], \label{eq:path_discrete2}
\end{align}
\end{subequations}
where $\ell(\vect{x}_k,\vect{u}_k)$ and $\ell_{\mathrm{f}}(\vect{x}_N)$ are the running and the terminal costs, \eqref{eq:dyn_discrete}  is the integrated discretized version of the reduced robot dynamics in \eqref{eq:simplified_2ropes_minimal}, 
\eqref{eq:path_discrete} and \eqref{eq:path_discrete2} are the \textit{path} constraints, and \eqref{eq:bound_discrete}  are the boundary conditions.
The control inputs include both the rope forces and the leg impulse $\vect{u}_k = \mat{f_{rl,k}, f_{rr,k}, \vect{f}_{\text{leg},k}} \in \Rnum^5$. 
Notice that $\vect{f}_{\text{leg}} \in \Rnum^3$ is applied only during a short thrusting phase ($t \leq t_{\text{th}}\in \Rnum$) at the beginning of the jump\footnote{The jump is inherently a hybrid motion planning problem due to contact switch at lift-off.
We make it time-variant by fixing the thrusting phase duration.}.
The states are the generalized variables of the reduced model introduced in Section \ref{sec:alpine} and their first derivatives:
\begin{equation}
	\vect{x}_k = \mat{\psi_k & l_{1,k} & l_{2,k} & \dot{\psi}_k & \dot{l}_{1,k} & \dot{l}_{2,k}   }^T \in \Rnum^6 .
\end{equation}
Consequently, the \textit{state-space} representation of the
nonlinear dynamics~\eqref{eq:simplified_2ropes_minimal}, casted in input-affine form \cite{alpine}, can be derived:
\begin{align}
    \begin{array}{ll}
      \dot{\vect{x}}_k = \vect{f}(\mathbf{x}_k) + \vect{g}(\vect{x}_k) \mathbf{u}_k,   \\
    \end{array}
\end{align}
where:
\begin{align}
	&\vect{f}(\vect{x}_k) =
	\mat{
		\mat{\dot{\psi}_k & \dot{l}_{1,k} & \dot{l}_{2,k}}^T\\
		\vect{A}_{d}(\vect{x}_k)^{-1} \left(\vect{g} - \vect{b}_{d}(\vect{x}_k)  \right)  
    },
  \\
  & \vect{g}(\vect{x}_k) = \mat{\vect{0}_{3 \times 1}	                           	& 				    \vect{0}_{3 \times 1}   & \vect{0}_{3 \times 3} \\
				                \vect{A}_{d}(\vect{x}_k)^{-1}\hat{\vect{a}}_{rl}  & \vect{A}_{d}(\vect{x}_k) ^{-1}\hat{\vect{a}}_{rr} & \frac{1}{m}  \vect{I}_{3 \times 3} }.
  \label{eq:DynamicsStateSpaceMinimal}
\end{align}
Finally, we also consider the jump duration $t_f\in \Rnum^{+}$ as a decision variable.
\par
To solve the optimization problem in~\eqref{eq:NLP}, we use a \textit{single shooting} approach eliminating the \textit{dependent} state variables $\vect{X}$ that can be expressed by the controls $\vect{U}$.
In particular, we compute the state vector $\vect{x}_{k+1}$ from the input $\vect{u}_k$, by integrating the dynamics via a Runge-Kutta 4 (RK4) method starting from the initial state $\vect{x}_0=\hat{\vect{x}}_0$ (Boundary Conditions) till the end of the horizon.
%

\noindent
{\bf State Constraints.}
State constraints are related to regions of the operation space that are not accessible.  A constraint is added to prevent collision with the wall mesh for the whole jump duration. 
Since the rocky wall can have an irregular shape, or there can be obstacles that the robot should overcome to reach its final destination, we model the inaccessible area as a region whose boundary is a surface. 
We assume that this surface can be expressed by a smooth 2D manifold expressed by
$Q(\vect{x}) = 0$.  
Therefore, the admissible region is generally given by $Q(\vect{x}) \geq 0$ and could be non-convex.
Considering a function $f_{\text{mesh}}(.)$ that interpolate the mesh at a point $(Y, Z)$, see Fig. \ref{fig:full_cost_map}(a), the constraints can be expressed as:
%
\begin{align}
 p_{X}(\vect{x}_k)\geq f_{\text{mesh}}(p_{Y}(\vect{x}_k), p_{Z}(\vect{x}_k)), \ \forall \ k \in \left[0, N\right]. 
 \label{eq:wall_constraint}
\end{align}
Additionally, we found it useful to encourage a clearance $h \in\Rnum$ including an inequality, for example, at the middle of the duration of the jump trajectory:
\begin{align}
	\vect{n}_c^T\vect{p}(\vect{x}_{N/2})    \geq h,
    \label{eq:viapoint_constraint}
\end{align}
where $\vect{n}_c$ is the surface normal of the terrain at contact location computed as:
\begin{align}
 \vect{n}_c = \frac{\nabla F}{\Vert \nabla F \Vert},
\end{align}
with $F =0$, the parametric function defining the terrain 2D curve in the space.
In our specific case, because of the definition of the inertial frame in Fig. \ref{fig:3dmodel2anchors_propellers}, we define the curve as a scalar field  on the $Y, Z$ plane $F=f_{mesh}(Y,Z) - X$:
\begin{align}
 \vect{n}_c = \frac{\mat{-1 & \frac{\partial f_{mesh}}{\partial y} & \frac{\partial f_{mesh}}{\partial z}}^T }{\sqrt{1+  \frac{\partial f_{mesh}}{\partial y}^2 + \frac{\partial f_{mesh}}{\partial z}^2}}.
 \end{align}
%
\noindent{\bf Landing Point Optimization.}
We improved the problem formulation from the original one found in~\cite{alpine}, by optimizing also the \textit{landing} location on the terrain mesh (i.e., height map/voxel map) for the intermediate jumps.
To achieve this, we constrain the solver to search the landing position inside the landing patch selected for the jump at hand \eqref{eq:landing_constraint_x}, while  \eqref{eq:landing_constraint_mesh} enforces consistency with the mesh:
\begin{subequations}\label{eq:landing_constraint}
\begin{align} 
    &\vect{p}_{c,i} -\vect{d}/2 \leq \vect{p}_{Y,Z}(\vect{x}_N) \leq \vect{p}_{c,i}+\vect{d}/2, \label{eq:landing_constraint_x}\\
    &\Vert p_X(\vect{x}_N) -  f_{\text{mesh}}(p_Y(\vect{x}_N), p_Z(\vect{x}_N)) \Vert \leq \epsilon,
    \label{eq:landing_constraint_mesh}
\end{align}
\end{subequations}
where $\vect{d}$ denotes the size of the $i$-th patch along the $Y$- and $Z$-directions, $\mathbf{p}_{c,i} \in \mathbb{R}^2$ denotes $Y, Z$ coordinates of the center location of the $i$-th patch, and $\epsilon \geq 0$ is a fixed relaxation parameter. The terrain cost associated with the landing location selected by the optimization will be accounted for in the objective \eqref{eq:cost}.
%
Since in the sequence of jumps dictated by the outer loop, the final jump must land at the desired goal location  $\vect{p}_f$, the landing constraints   \eqref{eq:landing_constraint} are replaced by: 
\begin{align}
  &\Vert \vect{p}(\vect{x}_N) -  \vect{p}_f \Vert \leq \epsilon.\label{eq:landing_constraint_last_jump}
 \end{align}
\noindent
{\bf Actuation Constraints.}
The forces that act along the direction of the respective rope axes $\hat{\vect{a}}_{rl}$, $\hat{\vect{a}}_{rr}$, see Section \ref{sec:alpine}, can not push the robot while the leg can not pull on the rock wall:
\begin{align}
 \vect{n}_c^T \vect{f}_{\text{leg},k} \geq 0, \quad \mathbf{f}_{r,k} \leq \mathbf{0} , \ \forall \ k \in \left[0, N \right),
 \label{eq:unilaterality}
\end{align}
with $\mathbf{f}_{r,k} = \left[f_{rl,k}, f_{rr,k} \right]^T$.
\par
Additionally, both the rope forces and the leg impulse are bounded by actuation limits and a second-order friction cone:
\begin{align}
   \Vert \vect{f}_{\leg,k} \Vert \leq f_{\leg, \max},& \ \forall \ k \in \left[0, N \right),  \\
   -\mathbf{f}_{r,k}^{\max} \preceq  \mathbf{f}_{r,k},& \ \forall \ k \in \left[0, N \right),\\
   \Vert \vect{t}_{x}^T \vect{f}_{\leg,k}  + \vect{t}_{y}^T \vect{f}_{\leg,k}  \Vert \leq \mu \vect{n}_c^T \vect{f}_{\leg,k},& \ \forall \ k \in \left[0, N \right),
  \label{eq:max_rope_force}
\end{align}
where $\mu$ is the \textit{friction coefficient}, and $\vect{t}_{x}$ and $\vect{t}_{y}$ the unit vectors perpendicular to the contact normal.
We observed that strongly reducing $\mathbf{f}_{r, max}$ encourages the robot to rely more on its leg and detach further from the wall, spending 
more time in the air and thereby giving the rope hoist more time to accelerate.
%

\noindent
{\bf Objective Function.}
We do not consider any particular terminal cost. 
For the running cost, we use a smoothing term for the rope forces, a second term that penalizes the hoist work (i.e., work made to wind/unwind the ropes), and a third term to minimize the terrain cost at the \textit{landing location}. 
Thus, the cost function is:
{\small
\begin{align}
  J & =  w_s \sum_{i \in \{r,l\}} \left[ \sum_{k=0}^{N-1} (\mathbf{f}_{{ri,k}} - \mathbf{f}_{{ri,k-1}})^2 + 
       w_{hw}\sum_{k=0}^{N-1} \vert \mathbf{f}_{{ri,k}} \dot{\mathbf{l}}_{i,k}  \vert T_s\right] + \nonumber\\      
      & + w_{l}c_l(\vect{p}(\vect{x}_N)),   
    \label{eq:cost}
\end{align}
}
where $c_l(.)$ is the function evaluating the cost at a certain location and $w_s$, $w_{hw}$, and $w_{l}$ are the weights associated with the three costs. 
%
%
The time $t_f$ is initialized with the time constant for the system linearized around the initial position $\vect{p}_0$. 
The rope forces are initialized with zeros, and the leg impulse with $\mathbf{f}_{\leg,\max}$.
%

\noindent
\textbf{Contribution to Outer Optimization Cost}.
From the output of the inner loop optimization, we compute the cumulative cost value $f_2$ \eqref{eq:fitness_term_2} from multiple jumps of: 1) the total energy consumption  $c_{e,j}$, 2) the punctual cost of the landing location evaluating the last point of the trajectory computed by the inner loop, and 3) the degree of convergence.
%
The total energy consumption for the j-th jump  will be the sum of the hoist work (first term) and the thrusting work (second term) due to the action of the thrusting force, which is the  kinetic energy at lift-off starting from standstill:
\begin{equation}
  c_{e,j} =  \sum_{i \in \{r,l\}} \sum_{k=0}^{N-1} \vert \mathbf{f}_{{ri,k}} \dot{\mathbf{l}}_{k}  \vert T_s  + 
   + \mathbb{I}_{t \in [0,t_{th}]}   \frac{1}{2}m\vect{\dot{p}}(\vect{x}_k)^T\vect{\dot{p}}(\vect{x}_k), 
\end{equation}
where $\mathbb{I}_{t \in [0, t_{th}]}$ is the indicator function selecting the samples $k$ such that $0 \leq t_k \leq t_{th}$.
%

\section{Experimental Results}\label{sec:results}
\begin{figure*}[!tbh]
   \centering
\begin{subfigure}{0.3\textwidth}
    \centering
    \includegraphics[width=\linewidth, trim={2.5cm 2.5cm 2cm 5cm}, clip=true]{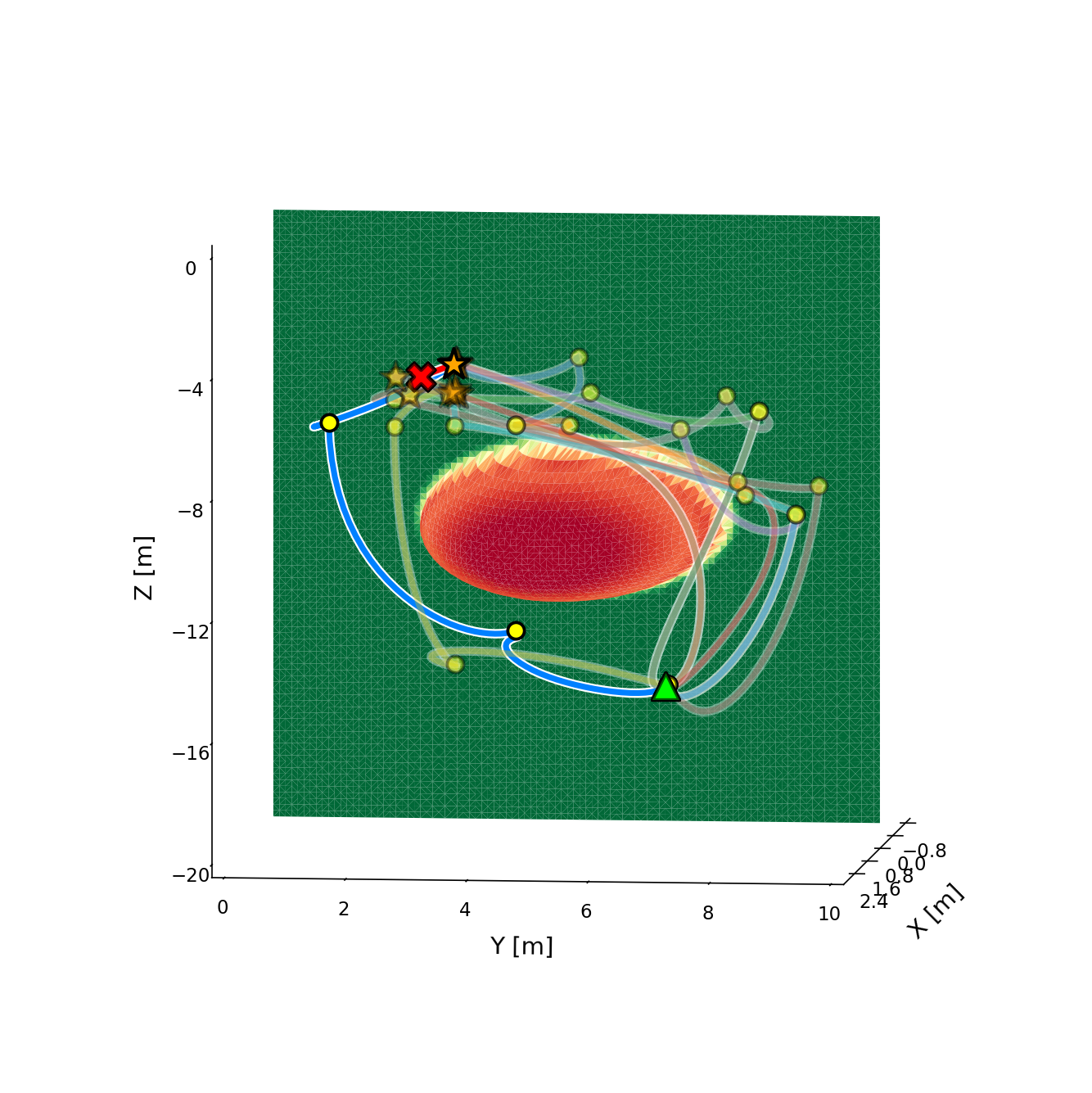}
    \caption{Hemispheric map.}
\end{subfigure}\hfill
\begin{subfigure}{0.3\textwidth}
    \centering
    \includegraphics[width=\linewidth, trim={2.5cm 2.5cm 2cm 5cm}, clip=true]{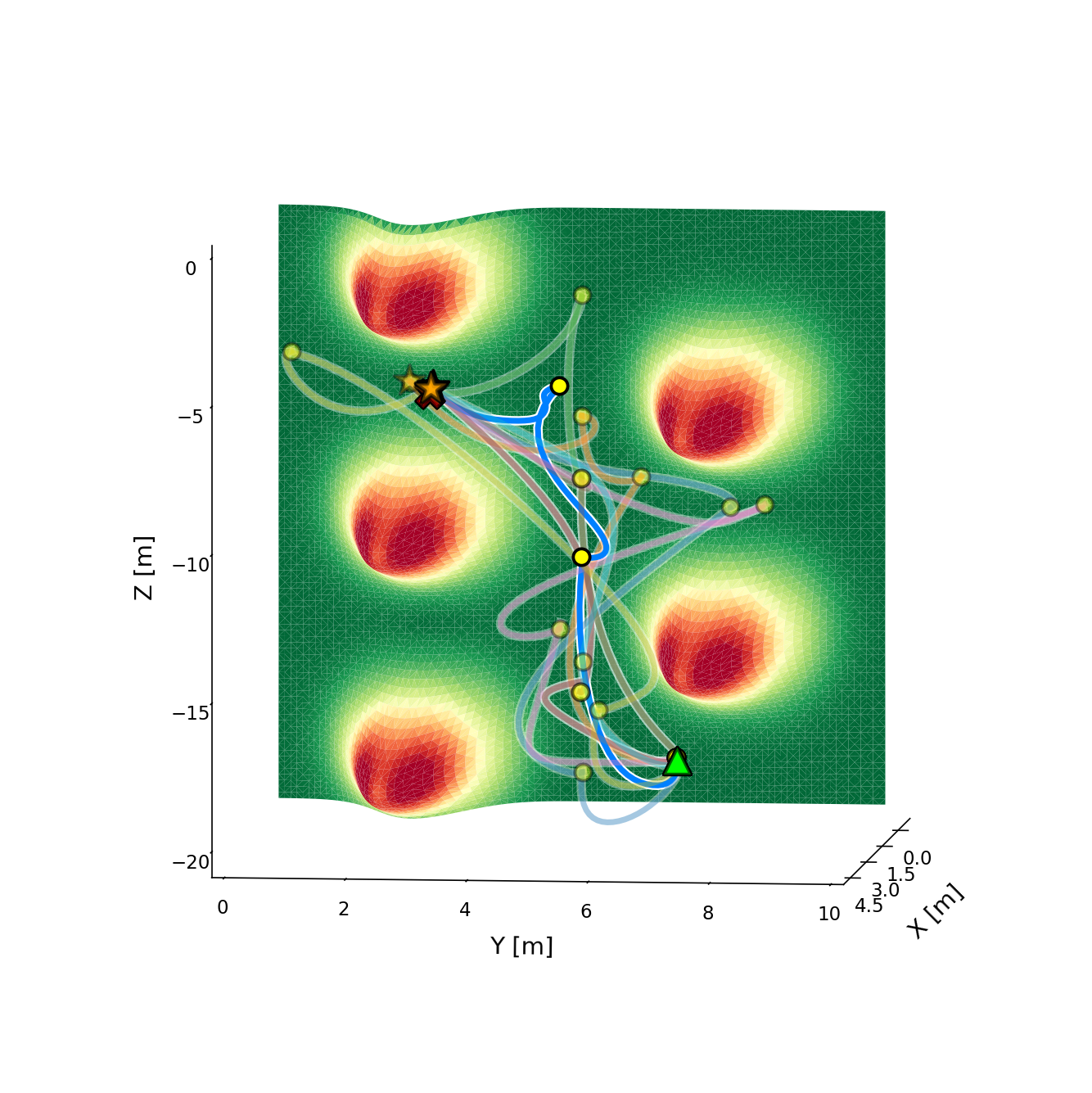}
    \caption{Bulged Pillar Wall map.}
\end{subfigure}\hfill
\begin{subfigure}{0.3\textwidth}
    \centering
    \includegraphics[width=\linewidth, trim={2.5cm 2.5cm 2cm 5cm}, clip=true]{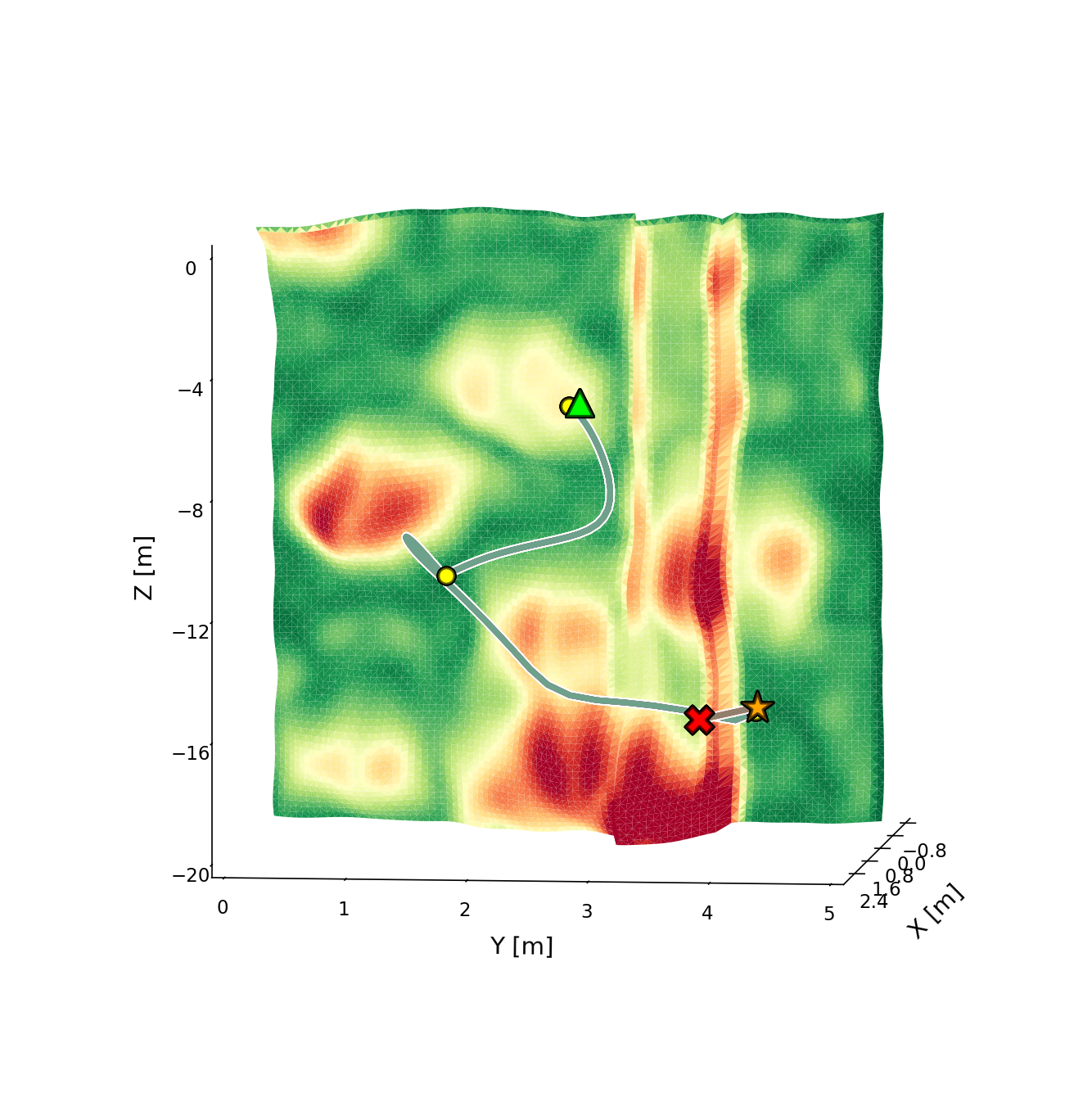}
    \caption{Rock-Wall Terrain.}
\end{subfigure}\hfill
\caption{\small Computed trajectory for three types of maps. With \startmarker\ is depicted the starting point, with \goalmarker\ the desired goal, with \jumpmarker\ the intermediate jumps, and with \achievedmarker\ the reached goal. Multiple feasible trajectories were generated using different random seeds; the optimal one is reported in blue.}
    \label{fig:terrain_results}
    \vspace{-0.5cm}
\end{figure*}
This section presents simulation results showing how the proposed framework enables robot navigation along a mountain wall.

We present 3 experiments, where we apply the optimization output to: 1) a big hemispheric obstacle to show the capability to split the jump into smaller intermediate jumps, 2) a bulged terrain to show the capability of our approach to avoid complex obstacles patterns, and 3) a rocky wall template where we track the optimization results with an \gls{mpc} controller within a realistic simulation environment.
Experiments were run on a notebook with a 20-core Intel Core i7 CPU (up to $4.9~\mathrm{GHz}$). As the proposed approach is intended for offline optimization, computation times are not reported. Improving computational efficiency is left for future work.

We modified the \gls{cem} algorithm to ensure every class (jump count and patch ID) has a minimum non-zero selection probability.
This acts as learning-rate regularization, preventing elite samples from dominating probability updates, thus maintaining exploration and avoiding premature convergence to loaca minima.
Additionally, biasing the first iteration by injecting elites along the straight start-to-goal path provides an effective heuristic for initial patch ID probabilities before the algorithm adapts to obstacles.

It should also be noted that since each evaluation within \gls{cem} is independent, they can be calculated in parallel to improve the total execution time of our framework.
The inner-loop optimization, originally in Matlab using \texttt{fmincon}, was ported to C++ (along with the \gls{ocp}) to improve planning and control performance.
The outer loop optimization is written in Python, leveraging multi-threading for parallelization, where each thread runs an independent instance of the internal loop combination and communicates only via shared summary statistics protected by locks.

Height maps ($10\times20~\mathrm{m}$, $0.5~\mathrm{m}$ resolution, $1~\mathrm{m}$ path spacing) were generated using Gaussian filters for pillar-like bulges.
In experiment 3, we added local ridges, dihedrals, and multiscale Perlin-like fractal noise to model multi-frequency surface roughness.
These maps are passed to the cost map filters (Section~\ref{sec:cost_map}) and bilinearly interpolated for smoothness.

\begin{figure}[!htb]
   \centering   
    
    
    \includegraphics[width=0.85\columnwidth]{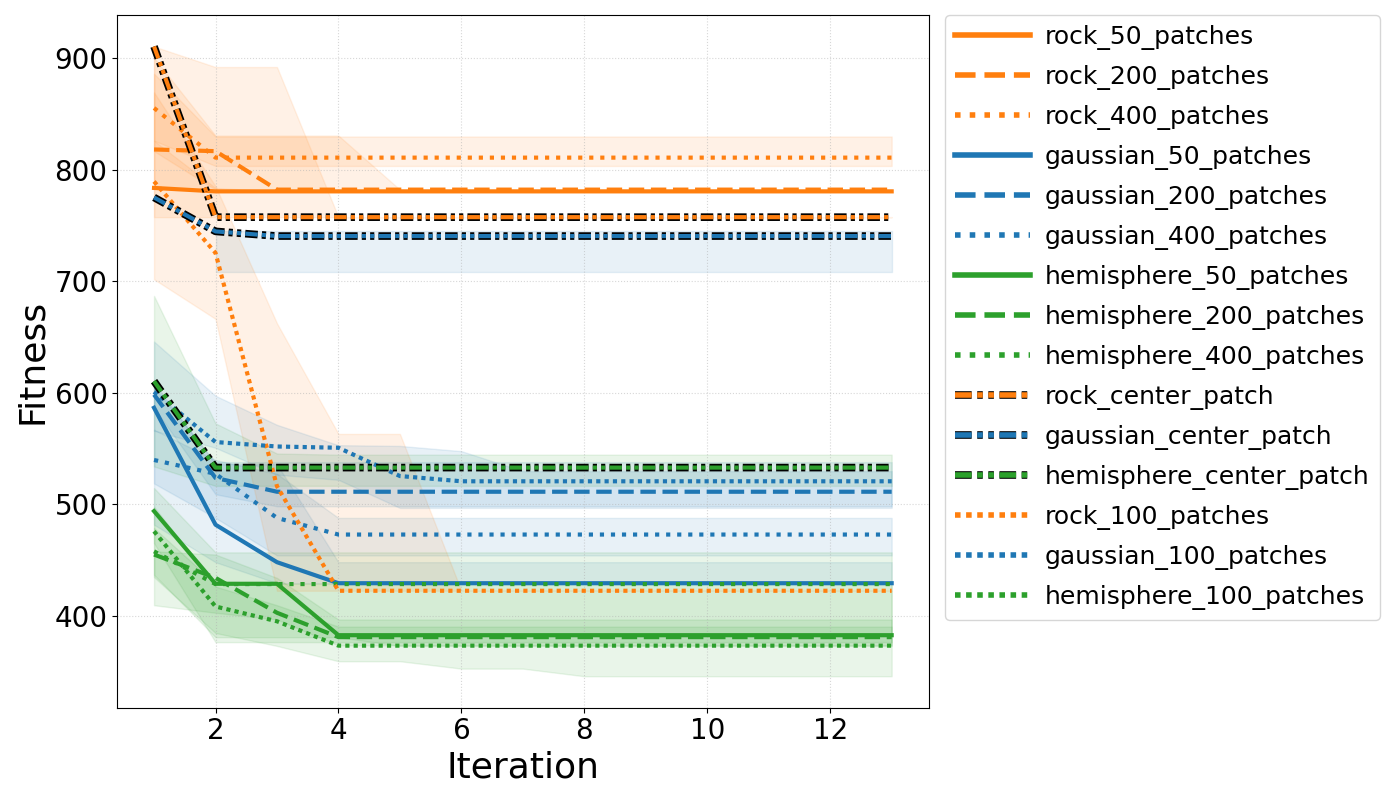}\vspace{0.2cm}
    \includegraphics[width=0.85\columnwidth]{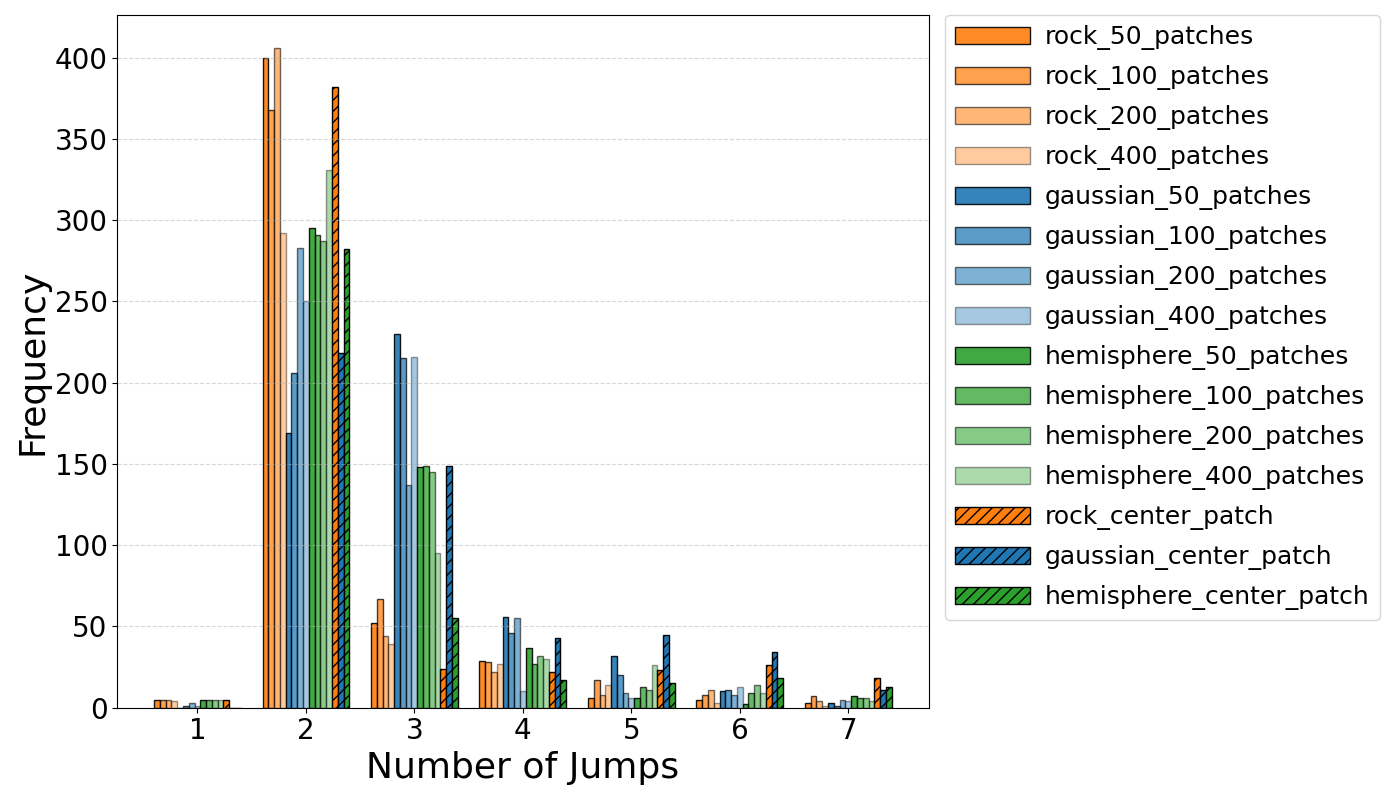}
    \caption{\small (Upper) Fitness evolution for the considered terrain models using different patch resolutions, together with the non-optimized reference obtained by fixing the contact point at the patch center ($0.5$). (Lower) Distribution of the selected number of jumps for the same configurations.}
    \label{fig:fitness_convergence_number_of_jumps}
    \vspace{-0.5cm}
\end{figure}

Parameters are listed in Tab.~\ref{tab:all_params}.
To assess reliability, we varied (see Fig.~\ref{fig:fitness_convergence_number_of_jumps}) the patch count (50, 100, 200, and 400), finding that fewer patches yield worse fitness and convergence.
Omitting the inner-patch optimization to select the best landing point produced the worst results, confirming that patches are non-uniform and optimal landing sites rarely lie at the center. 
The source code is available at: \url{https://github.com/mfocchi/climbing_robots2/tree/multi-jump}
\begin{table}[t]
    \centering
    \caption{Hyper-Parameters}
    \resizebox{\columnwidth}{!}{%
    \begin{tabular}{l c c}
        \hline
        \textbf{Parameter} & \textbf{Symbol} & \textbf{Value} \\
        \hline\hline

        \multicolumn{3}{c}{\textbf{Physical and Simulation Parameters}} \\
        \hline

        Robot mass [kg]                          & $m$                        & 5 \\
        Gravity acceleration [m/s$^2$]           & $g$                        & 9.81 \\
        Left anchor position [m]                 & $\vect{p}_{a,1}$           & $\mat{0 & 0 & 0}$ \\
        Right anchor position [m]                & $\vect{p}_{a,2}$           & $\mat{0 & 10 & 0}$ \\
        Friction coefficient [-]                 & $\mu$                      & 0.8 \\

        \hline\hline
        \multicolumn{3}{c}{\textbf{Inner-loop Optimization Parameters}} \\
        \hline

        Integration method                       & --                         & RK4 \\
        Dynamic discretisation steps             & $N_{\text{dyn}}$           & 30 \\
        Sub-integration steps                    & $N_{\text{sub}}$           & 5 \\
        Thrust impulse duration [s]              & $T_{\text{th}}$            & 0.05 \\

        Max. leg force [N]                       & $f_{\text{leg}}^{\max}$    & 150 \\
        Min. Max. rope force [N]                      & $f_{r}^{\min}$/$f_{r}^{\max}$             & 15/60 \\
        Inner loop. cost weights                        & $\mat{w_{s}&w_{hw}& w_l }$                     &$\mat{ 0.001 &0& 1000}$ \\
        Jump clearance [m]                       & $h$                        & 0.5 \\

        \hline\hline
        \multicolumn{3}{c}{\textbf{Cost Function Weights}} \\
        \hline

        Non-convergence penalty                  & $w_p$                      & $1 \times 10^{7}$ \\
        Energy consumption weight                & $w_e$                      & 1 \\
        Average costmap patch weight\textbf{(N/a)}             & $w_{cp}$                   & 0 \\
        Landing costmap weight                   & $w_{lc}$                   & 100 \\
        trajectory length weight                     & $w_{lt}$                   & 20 \\

        \hline\hline
        \multicolumn{3}{c}{\textbf{Terrain and Environment Modeling}} \\
        \hline

        Terrain grid resolution                  & $N_g$                      & 100 \\
        Wall depth [m]                           & $d_w$                      & 10 \\
        Maximum ridge depth [m]                  & $d_r^{\max}$               & 0.5 \\
        Terrain size (vertical/lateral) [m]            & $L_z$/ $L_y$                    & -10/20 \\
        Patch grid width/height                         & $N_w$/$N_h$                        & 10/20 \\
        Number of terrain patches                & $N_p^{\text{patch}}$       & 200 \\
        Patch width/height [m]                          & $\Delta y$/$\Delta z$                      & derived \\

        \hline\hline
        \multicolumn{3}{c}{\textbf{Costmap Filtering Weights}} \\
        \hline
        First-derivative (gradient) weight       & $w_{sl}$                & 1 \\
        Second-derivative (Laplacian) weight     & $w_{sd}$                & 1 \\

        \hline\hline
        \multicolumn{3}{c}{\textbf{\gls{cem} Hyper-parameters}} \\
        \hline

        Number of threads                        & $N_t$                      & 10 \\

        Maximum number of jumps                  & $n_{\max}$                 & 6 \\
        Discrete \gls{cem} dimension                   & $d_d$                      & $J_{\max}+1$ \\

        Population size                          & $N_{\text{pop}}$           & 200 \\
        Percentage of elites                        & $N_e$                      & 20 \\
        Maximum iterations                      & $K$                        & 50 \\
        Smoothing factor                         & $\alpha$                   & 0.5 \\
        Minimum discrete probability             & $p_{\min}$                 & $0.8/N_p$ \\

        \hline
    \end{tabular}}
    \label{tab:all_params}
    \vspace{-0.5cm}
\end{table}

%
%


\subsection{Use Case 1: Hemispheric Obstacle }
We first evaluate our framework in a scenario where the robot must find a feasible path from the start to the goal via multiple intermediate jumps. This constraint arises either from physical hardware limitations or from obstacles that prevent a single large jump.
In this setup, a hemispherical obstacle is positioned between the robot’s initial location and the goal above. 
The goal is located above the start, requiring the robot to perform upward jumps that are more likely to hit the actuation bounds because there is no aid from gravity.
Because the robot cannot clear the obstacle in a single \textit{upward} jump, it must perform a sequence of jumps to navigate around it, see Fig.~\ref{fig:terrain_results} (left).
As can be seen in Fig. \ref{fig:fitness_convergence_number_of_jumps} (lower plot), the \gls{cem} solver explores several jumps and the fitness converges to a value of \textit{321.3}, see Fig.~\ref{fig:fitness_convergence_number_of_jumps} (upper plot), after 4 iterations.

\subsection{Use Case 2: Multi Bulge Terrain}
In this use case, we test the effectiveness of our approach in producing feasible obstacle-free plans. The goal in the following use case is located below the start (see Fig. \ref{fig:terrain_results}(center), the robot is initialized at the $YZ$ starting pose $\mathbf{p}_0$=$( 6.556, -18.49) \ m$ and goal pose  $\mathbf{p}_f$=$(2.47, -6.3) \ m$, so the robot will perform downward jumps. In this way, we would expect the robot to perform fewer jumps thanks to the help of gravity, but the presence of a complex pattern of obstacles on the way requires at least 3 jumps. The total duration of the multi-jump sequence is 9.45s.
\par
%

\subsection{Use Case 3: Rocky terrain Gazebo Simulation}
\label{sec:gazebo_sim}
Finally, we evaluate the effectiveness of our approach on a realistic rocky wall by simulating the robot with an \gls{mpc} controller that tracks the optimized multi-jump sequence in a 
Gazebo environment, see Fig.~\ref{fig:gazebo_simulation}. The simulation computes the full constrained robot dynamics using the ODE physics engine~\cite{OdePhysicsEngine}. Visualizations of the resulting behaviors are provided in the supplementary video link \href{https://youtu.be/yUsPvjpKSzw}{\textbf{https://youtu.be/yUsPvjpKSzw}}. 
We employ the URDF formalism to describe the robot model and the Pinocchio library~\cite{carpentier2019pinocchio} for kinematics and dynamics computation.
\par
For a $11.18 \ m$ jump, from $\vect{p}_0 = (2.5, -6) \ m$ to $\vect{p}_f=(4.5, -17) \ m$, we deliberately set actuation constraints $f_{\leg}^{\text{max}}=150 \ N$ and $f_r^{\text{max}} = 60 \ N$ such that the robot cannot reach $\vect{p}_f$ with a single  feasible jump. 
Given the mesh of the rocky wall, we solve the bi-level optimization problem to find the optimal landing locations for the intermediate jumps. Rather than strictly following the corresponding planned trajectories, which could lead to error accumulation due to imperfect controller tracking, we re-optimize each jump after landing, using the robot’s actual landing location as the new initial condition.
Each optimization provides the values of the initial leg impulse, the pattern of the rope forces, and the jump duration for each jump. 
%
The simulation runs at $1~kHz$, 
since the optimized trajectory has a different time discretization ($d_t = t_f / N$), depending on the jump duration $t_f$, appropriate interpolations are performed to adapt the rate difference\footnote{Integration errors are mitigated performing multiple sub-integration steps between consecutive knots during the rollout. While physical consistency may degrade for longer jumps, this issue can be addressed by increasing the number of knots according to the desired jump length.}.   
Each jump is regulated by a state machine that sequences a set of phases, i.e., \emph{leg reorient}, \emph{thrusting}, \emph{flying}, and \emph{landing}. 
During the thrusting phase, we generate the pushing impulse $\vect{f}_{\leg}$ at the \gls{com} by applying a force $\vect{f}_c$ at the contact point.
Since $\vect{f}_{leg}$ is applied at the pushing leg tip, 
to avoid generating moments at the \gls{com}, we perform a preliminary orientation of the prismatic leg to align it in the direction of $\vect{f}_{leg}$.
During the flying phase, the rope (prismatic) joints are actuated in \textit{force} control mode. 
After liftoff, the leg no longer influences the linear motion of the base; therefore, during the flight phase, an \gls{mpc} action is superimposed on the optimized force profiles to reject disturbances and minimize the tracking error w.r.t. the reference trajectory during flight~\cite{alpine}.
Since the robot has no control authority in the direction perpendicular to the rope plane while airborne, the controller also commands the thrust $f_p$ of a  propeller mounted on the robot back (see Fig.~\ref{fig:3dmodel2anchors_propellers}) to regulate the  $\psi$ state~\cite{alpine}. The propeller generates a force $\vect{\hat{x}}_b f_p$ along the base-link $X$ unit axis $\vect{\hat{x}}_b \in \Rnum^3$, which is orthogonal to the rope plane.
For control computation, we use a slightly modified reduced-order model in which the term $\vect{\hat{x}}_b f_p$ is added to the right-hand side of \eqref{eq:newton}. 
%
%
After each intermediate jump, the robot's velocity is assumed to be completely dissipated by an impedance controller on the prismatic leg.
 %
%
%
%


\section{Conclusions}
\label{sec:conclusion}
We presented a motion and contact planner based on bi-level optimization to navigate complex terrain environments, such as mountainous regions. 
The outer optimization layer relies on the \gls{cem}, while the inner loop formulates an \gls{ocp} to autonomously compute contact schedules, contact locations, and motion trajectories from a given terrain representation.
We extend the original problem formulation introduced in~\cite{alpine} by supporting arbitrary terrain meshes and by optimizing the landing location using a terrain cost map.
The planner is coupled with an \gls{mpc}-based controller and validated in simulation using the \emph{ALPINE} robot.
Current limitations of the framework include the treatment of cables, whose potential penetration into obstacles is not yet considered, as well as the absence of a fitness term accounting for the depletion level of the gas bottle supplying the pneumatic piston that actuates the leg. 
%
%
%
We plan to consider these aspects in future works, eg., finding landing locations to minimize obstacle penetration, together with experiments on hardware.
%

\renewcommand{\bibfont}{\small}
\printbibliography

@INPROCEEDINGS{focchi23icra,
  author={Focchi, Michele and Bensaadallah, Mohamed and Frego, Marco and Peer, Angelika and Fontanelli, Daniele and Del Prete, Andrea and Palopoli, Luigi},
  booktitle={IEEE International Conference on Robotics and Automation}, 
  title={{CLIO: a Novel Robotic Solution for Exploration and Rescue Missions in Hostile Mountain Environments}}, 
  year={2023},
  volume={},
  number={},
  pages={7742-7748}
}

@misc{OdePhysicsEngine,
  author       = {{Russell Smith}},
  title        = {Open Dynamics Engine},
  howpublished = {\url{https://www.ode.org/}},
  year         = {2001},
  note         = {Accessed: 2024-06-17}
}

@inproceedings{carpentier2019pinocchio,
	author = {Carpentier, Justin and Saurel, Guilhem and Buondonno, Gabriele and Mirabel, Joseph and Lamiraux, Florent and Stasse, Olivier and Mansard, Nicolas},
	booktitle = {IEEE International Symposium on System Integrations},
	title = {{The Pinocchio C++ library -- A fast and flexible implementation of rigid body dynamics algorithms and their analytical derivatives}},
	year = {2019}
}

@article{alpine,
title = {ALPINE: A climbing robot for operations in mountain environments},
journal = {Robotics and Autonomous Systems},
volume = {190},
year = {2025},
author = {Michele Focchi and Andrea Del Prete and Daniele Fontanelli and Marco Frego and Angelika Peer and Luigi Palopoli}
}

@article{kim2025contact-implicit,
author = {Gijeong Kim and Dongyun Kang and Joon-Ha Kim and Seungwoo Hong and Hae-Won Park},
title ={Contact-implicit Model Predictive Control: Controlling diverse quadruped motions without pre-planned contact modes or trajectories},
journal = {The International Journal of Robotics Research},
volume = {44},
number = {3},
pages = {486-510},
year = {2025}
}

@article{Mordatch2012DiscoveryOC,
  title={Discovery of complex behaviors through contact-invariant optimization},
  author={Mordatch, Igor and Todorov, Emanuel and Popovic, Zoran},
  journal={ACM Transactions on Graphics},
  year={2012},
  volume={31},
  pages={1 - 8}
}

@inproceedings{marconi2012sherpa,
  title={The SHERPA project: Smart collaboration between humans and ground-aerial robots for improving rescuing activities in alpine environments},
  author={Marconi, Lorenzo and Melchiorri, Claudio and Beetz, Michael and Pangercic, Dejan and Siegwart, Roland and others},
  booktitle={IEEE international symposium on safety, security, and rescue robotics},
  pages={1--4},
  year={2012}
}

@article{miki2022learning,
  title={Learning robust perceptive locomotion for quadrupedal robots in the wild},
  author={Miki, Takahiro and Lee, Joonho and Hwangbo, Jemin and Wellhausen, Lorenz and Koltun, Vladlen and Hutter, Marco},
  journal={Science robotics},
  volume={7},
  number={62},
  year={2022}
}

@article{uckert2020investigating,
  title={Investigating habitability with an integrated rock-climbing robot and astrobiology instrument suite},
  author={Uckert, Kyle and Parness, Aaron and Chanover, Nancy and Eshelman, Evan J and Abcouwer, Neil and Nash, Jeremy and Detry, Renaud and Fuller, Christine and Voelz, David and Hull, Robert and others},
  journal={Astrobiology},
  volume={20},
  number={12},
  pages={1427--1449},
  year={2020}
}

@article{cepolina2006roboclimber,
  title={Roboclimber versus landslides: design and realization of a heavy-duty robot for teleoperated consolidation of rocky walls},
  author={Cepolina, Francesco and Moronti, Marco and Sanguinet, Michelangelo and Zoppi, Matteo and Molfino, Rezia M},
  journal={IEEE Robotics \& Automation Magazine},
  volume={13},
  number={1},
  pages={23--31},
  year={2006}
}

@INPROCEEDINGS{hoffman21,
  author={Hoffman, Enrico Mingo and Parigi Polverini, Matteo and Laurenzi, Arturo and Tsagarakis, Nikos G.},
  booktitle={IEEE International Conference on Robotics and Automation}, 
  title={Modeling and Optimal Control for Rope-Assisted Rappelling Maneuvers}, 
  year={2021},
  volume={},
  number={},
  pages={9826-9832}
}

@article{bretl2006motion,
  title={Motion planning of multi-limbed robots subject to equilibrium constraints: The free-climbing robot problem},
  author={Bretl, Timothy},
  journal={The International Journal of Robotics Research},
  volume={25},
  number={4},
pages={317--342},
  year={2006}
}

@article{choi2023examining,
  title={Examining the mechanics of rope bending over a three-dimensional edge in ascending robots},
  author={Choi, Myeongjin and Ahn, Sahoon and Kim, Hwa Soo and Seo, Taewon},
  journal={Scientific Reports},
  volume={13},
  number={1},
  year={2023},
  publisher={Nature Publishing}
}

@INPROCEEDINGS{Tsikelis25,
  author={Tsikelis, Ioannis and Tsiatsianas, Evangelos and Kiourt, Chairi and Ivaldi, Serena and Chatzilygeroudis, Konstantinos and Hoffman, Enrico Mingo},
  booktitle={IEEE-RAS International Conference on Humanoid Robots}, 
  title={{AHMP}: Agile Humanoid Motion Planning with Contact Sequence Discovery}, 
  year={2025},
  volume={},
  number={},
  pages={33-40}
}

@INPROCEEDINGS{dhedin2025simultaneous,
  author={Dhédin, Victor and Zhao, Haizhou and Khadiv, Majid},
  booktitle={IEEE-RAS International Conference on Humanoid Robots}, 
  title={Simultaneous Contact Sequence and Patch Planning for Dynamic Locomotion}, 
  year={2025},
  volume={},
  number={},
  pages={245-252}
  }

@ARTICLE{Manchester24,
  author={Le Cleac'h, Simon and Howell, Taylor A. and Yang, Shuo and Lee, Chi-Yen and Zhang, John and Bishop, Arun and Schwager, Mac and Manchester, Zachary},
  journal={IEEE Transactions on Robotics}, 
  title={Fast Contact-Implicit Model Predictive Control}, 
  year={2024},
  volume={40},
  number={},
  pages={1617-1629}
}

@INPROCEEDINGS{Cauligi20,
  author={Cauligi, Abhishek and Culbertson, Preston and Stellato, Bartolomeo and Bertsimas, Dimitris and Schwager, Mac and Pavone, Marco},
  booktitle={IEEE Conference on Decision and Control}, 
  title={Learning Mixed-Integer Convex Optimization Strategies for Robot Planning and Control}, 
  year={2020},
  volume={},
  number={},
  pages={1698-1705}
}

@inproceedings{rubinstein1999cross,
  author = {Rubinstein, Reuven Y},
  title = {The Cross-Entropy Method for Combinatorial and Continuous Optimization},
  booktitle={Methodology And Computing In Applied Probability},
  volume={1},
  pages={127--190},
  year = {1999},
}

@article{Marcucci24,
author = {Marcucci, Tobia and Umenberger, Jack and Parrilo, Pablo and Tedrake, Russ},
title = {Shortest Paths in Graphs of Convex Sets},
journal = {SIAM Journal on Optimization},
volume = {34},
number = {1},
pages = {507-532},
year = {2024}
}


\end{document}